\documentclass[10pt,twocolumn,letterpaper]{article}

\usepackage[pagenumbers]{iccv} % To force page numbers, e.g. for an arXiv version

\definecolor{iccvblue}{rgb}{0.21,0.49,0.74}
\usepackage[pagebackref,breaklinks,colorlinks,allcolors=iccvblue]{hyperref}

\usepackage{multirow}
\usepackage{amsmath}
\usepackage{amssymb}
\usepackage{float}
\usepackage{bbding}

\def\paperID{8244} % *** Enter the Paper ID here
\def\confName{ICCV}
\def\confYear{2025}

\title{RAG-Adapter: A Plug-and-Play RAG-enhanced Framework \\ for Long Video Understanding}

\author{Xichen Tan\\
College of Computer Science and Technology, \\ National University of Defense Technology\\
Changsha, China\\
{\tt\small tanxc23@nudt.edu.cn}
\and
Yunfan Ye\\
School of Design, \\Hunan University\\
Changsha, China\\
\and
Yuanjing Luo\\
College of Computer and Mathematics, \\Central South University of Forestry and Technology\\
Changsha, China\\
\and
Qian Wan\\
Faculty of Artificial Intelligence in Education, \\ 
Central China Normal University\\
Wuhan, China\\
\and
Fang Liu\\
School of Design, \\
Hunan University\\
Changsha, China\\
\and
Zhiping Cai\\
College of Computer Science and Technology, \\
National University of Defense Technology\\
Changsha, China\\
}

\begin{document}
\maketitle
\begin{abstract}
Multi-modal Large Language Models (MLLMs) capable of video understanding are advancing rapidly. To effectively assess their video comprehension capabilities, long video understanding benchmarks, such as Video-MME and MLVU, are proposed. However, these benchmarks directly use uniform frame sampling for testing, which results in significant information loss and affects the accuracy of the evaluations in reflecting the true abilities of MLLMs. To address this, we propose \textbf{RAG-Adapter}, a plug-and-play framework that reduces information loss during testing by sampling frames most relevant to the given question. Additionally, we introduce a \textbf{Grouped-supervised Contrastive Learning (GCL)} method to further enhance RAG-Adapter’s sampling effectiveness through fine-tuning on our constructed \textbf{MMAT} dataset. 
% Using RAG-Adapter’s sampling results, we define two metrics — Average Similarity Score (ASS) and Necessary Information Frame (NIF) — to evaluate the construction quality and complexity of the benchmarks. 
Finally, we test numerous baseline MLLMs on various video understanding benchmarks, finding that RAG-Adapter sampling consistently outperforms uniform sampling (e.g., GPT-4o's accuracy increases by 9.3\% on Video-MME), providing a more accurate testing method for long video benchmarks.

\end{abstract}    
\section{Introduction}
\label{sec:intro}

In the field of video understanding, research on short videos progresses earlier and more extensively than on long videos, primarily due to the quadratic complexity constraint of transformer-based models in handling long sequences. To mitigate this, many long video models, such as MovieChat~\cite{song2024moviechat} and LlamaVid~\cite{li2025llama}, introduce input token compression algorithms to reduce computational costs.

To evaluate the long video understanding capabilities of MLLMs, several specialized long video benchmarks have been proposed, including Video-MME~\cite{fu2024video} and MLVU~\cite{zhou2024mlvu}. However, these benchmarks do not standardize the number of input frames during testing due to variations in models’ maximum frame capacities. Moreover, not all MLLMs support one-frame-per-second sampling (assumed sufficient to capture content). For these models, testing relies on uniformly sampled frame subsets. In Video-MME, for instance, the longest test video spans one hour, yet only four uniformly sampled frames are used at minimum, often omitting critical information. This leads to responses resembling random guesses and makes it challenging to accurately evaluate true model performance.

\begin{figure}[t]
    \vskip -0.1in
    \begin{center}
    \centerline{\includegraphics[width=1\linewidth]{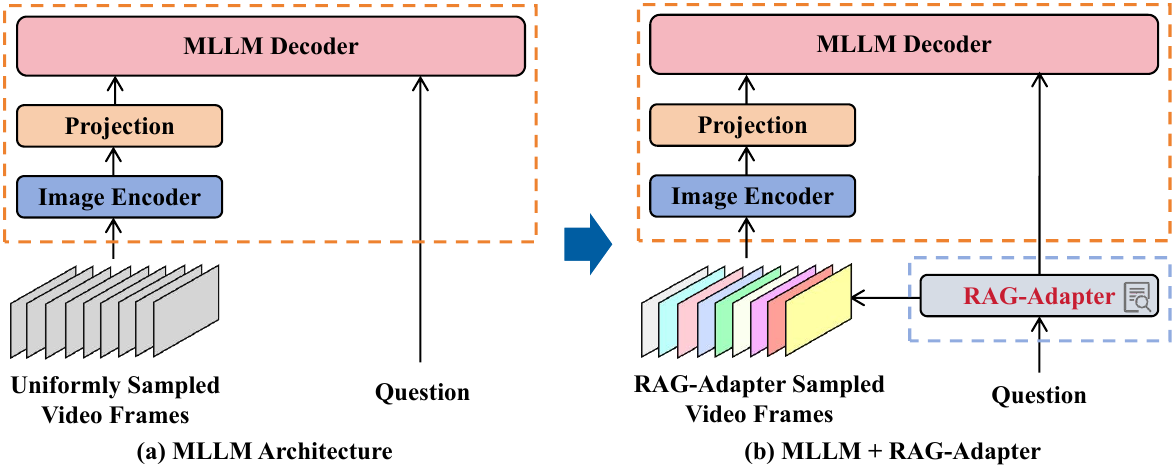}}
         \caption{(a) and (b) show a comparison between scenarios with and without the RAG-Adapter framework, respectively.}
     \label{teaser}
     \end{center}
     \vskip -0.5in
\end{figure}

To address the testing challenges in existing long video benchmarks, we propose a plug-and-play RAG-enhanced (Retrieval Augmented Generation) optimization framework, \textbf{RAG-Adapter}. As illustrated in~\Cref{teaser}, RAG-Adapter operates without modifying the internal architecture of MLLMs, instead focusing on video frame input. By retrieving the Top\(K\) most relevant video frames, it replaces the uniform sampling method, significantly reducing information loss. This straightforward yet effective approach more accurately evaluates the true long video understanding capabilities of MLLMs.

Although the approach is straightforward, research directly integrating RAG with MLLMs is limited. This is mainly because RAG-Adapter's retrieval performance depends heavily on similarity matching between embeddings generated by its text and image encoders (~\Cref{pipeline}). The embeddings produced by open-source encoders may be suboptimal for long video understanding tasks. Therefore, we fine-tune these encoders through contrastive learning to better align similar embeddings, thereby enhancing the retrieval effectiveness of RAG-Adapter.

Given the challenge of directly locating relevant frames in long videos, we further construct a fine-tuning dataset, \textbf{MMAT}, using short video understanding benchmarks. We extract video frames and pair them with corresponding questions to create positive pairs for fine-tuning.

Additionally, as a single video may correspond to multiple questions, the Self-supervised Contrastive Learning (SCL) assumption that treats other questions as negative samples may mislead the model during training. To address this, we propose \textbf{Grouped-supervised Contrastive Learning (GCL)}, where all positive pairs involving the same video's frame share a common group label. GCL enables clearer differentiation between intra-group and inter-group embeddings, thereby enhancing RAG-Adapter's retrieval capabilities for video understanding tasks.

Using retrieval results from RAG-Adapter fine-tuned with GCL (RAG-Adapter, unless specified otherwise, is GCL fine-tuned), we introduce two metrics: \textbf{Average Similarity Score (ASS)} and \textbf{Necessary Information Frame (NIF)}. ASS measures the average similarity between the Top\(K\) frames retrieved by RAG-Adapter and the corresponding questions, while NIF represents the average minimum number of frames containing essential information needed to answer each question. The NIF reveals that, even for long video understanding benchmarks, a small subset of frames typically contains the required information, validating our approach of using a fixed number of frames (Top\(K\)) across models for fair evaluation.

Notably, the ASS and NIF metrics offered by RAG-Adapter serve as important indicators for evaluating benchmark quality. A lower ASS may indicate insufficient relevance between video content and questions, suggesting potential flaws in question formulation, while a lower NIF implies that fewer frames are needed, indicating lower question complexity. 

In summary, the main contributions of this work are:

1) We propose \textbf{RAG-Adapter}, a plug-and-play enhancement framework for MLLMs. By supplying input-level video frames relevant to test questions, RAG-Adapter enhances the video understanding capabilities of MLLMs without structural modifications.

2) We construct the \textbf{MMAT} fine-tuning dataset and propose \textbf{Grouped-supervised Contrastive Learning (GCL)} for long video understanding scenarios, enhancing RAG-Adapter's retrieval performance.

3) We introduce two metrics through RAG-Adapter: \textbf{Average Similarity Score (ASS)} and \textbf{Necessary Information Frame (NIF)}, as standards for evaluating benchmark quality and complexity in long video understanding. NIF further confirms that RAG-Adapter provides information that is both sufficient and effective.

4) Extensive experiments on open-source long video understanding benchmarks demonstrate the effectiveness of RAG-Adapter in enhancing the video understanding capabilities of existing MLLMs.
\section{Related Work}
\label{sec:related work}

\subsection{Multi-model LLMs~(MLLMs)}
MLLMs extend traditional LLMs by incorporating a visual encoder and projection layer, enabling image and video understanding. Video-based MLLMs~\cite{alayrac2022flamingo, yang2023vid2seq, ataallah2024minigpt4, maaz2023video, weng2024longvlm, he2024ma, shen2024longvu}, process sampled video frames as input, is essentially equivalent to image-based MLLMs~\cite{liu2024llavanext, li2023otter, chen2024far, bai2023qwen, glm2024chatglm} that support multiple images, even if not explicitly trained on video data. To handle more frames for long video understanding, many MLLMs reduce computational complexity by compressing the number of visual tokens at the input level. 

MovieChat~\cite{song2024moviechat} applies ToMe~\cite{bolya2022token} methods to merge similar tokens between adjacent frames. LLaMa-VID~\cite{li2025llama} reduces image tokens through average pooling, while Chat-UniVi~\cite{jin2024chat} uses the k-nearest-neighbor based density peaks clustering algorithm (DPC-KNN) to segment videos into events, and group tokens of each frame within these events. 

Although these models can support inputs of up to thousands of video frames, the NIF metric in~\Cref{nie} indicates that the relevant information needed to answer questions resides in only a small subset of frames. Furthermore, ablation experiments in~\Cref{experiment3} show that using more uniformly sampled frames can yield inferior performance compared to using only frames directly relevant to the questions.

\subsection{Long Video Understanding Benchmarks}

To evaluate MLLMs's long video understanding capabilities, several benchmarks have been proposed, including Video-MME~\cite{fu2024video}, and MLVU~\cite{zhou2024mlvu}. These benchmarks contain numerous manually annotated Q\&A pairs, with average video lengths exceeding 10 minutes. The video content covers a wide range of domains, spanning domains such as daily life, art, sports, and television. They comprehensively assess MLLMs's abilities in cognition, reasoning, summarization, and other aspects of long video comprehension.

Although these benchmarks provide a comprehensive evaluation of different aspects, during the testing phase, a uniform sampling of video frames is used for all questions. Clearly, the information required for each question varies, and there is a high likelihood that the relevant information may not be included in the uniformly sampled frames. Therefore, assessing the long video understanding capabilities of MLLMs in this manner is not entirely reasonable.

\begin{figure*}[t]
    \vskip -0.1in
    \begin{center}
    \centerline{\includegraphics[width=0.90\linewidth]{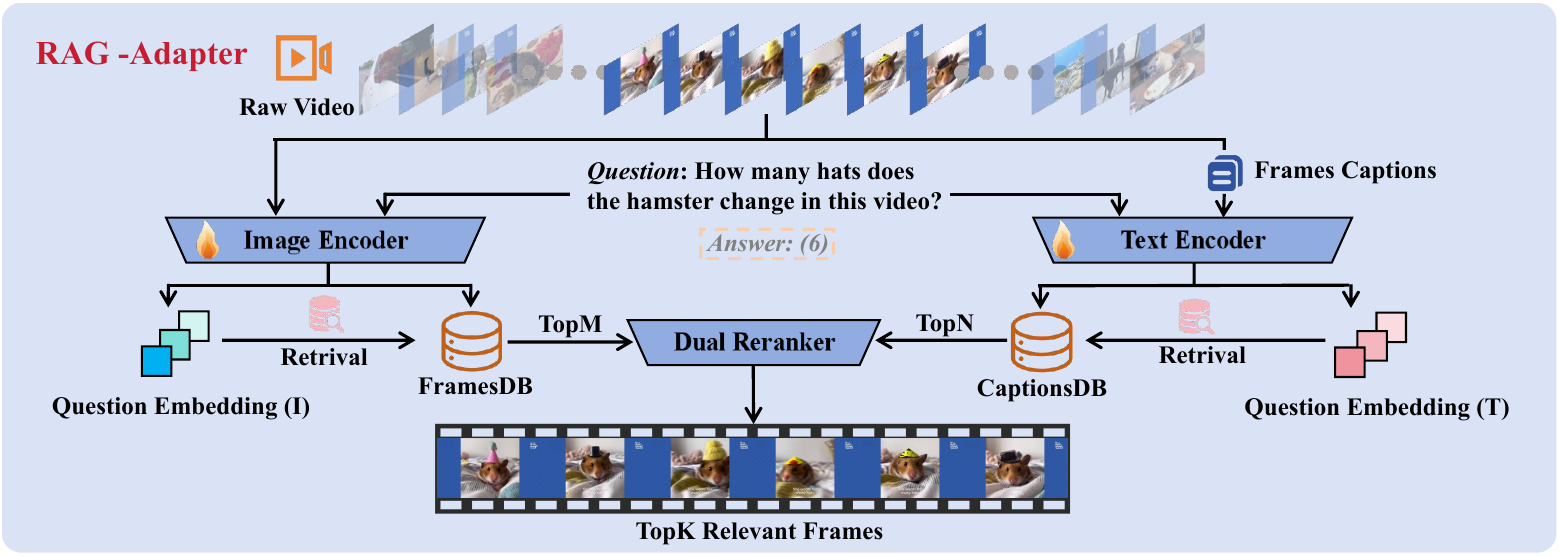}}
         \caption{The RAG-Adapter pipeline framework. Given a video and a question, the video frames and corresponding captions are encoded separately using image and text encoders and stored in databases. The question is encoded and retrieved using the same encoders. The Dual Reranker module selects the Top\(K\) frames relevant to the question. Details are provided in~\Cref{sec:pipeline}. To improve retrieval performance, both encoders are fine-tuned using Grouped-supervised Contrastive Learning (GCL), as described in~\Cref{sec:ft}.}
     \label{pipeline}
     \end{center}
     \vskip -0.3in
\end{figure*}

\subsection{Retrieval Augmented Generation~(RAG)}

RAG~\cite{lewis2020retrieval} was first introduced in NLP for retrieval augmentation, and rapidly inspired advancements in text retrieval, with optimizations targeting various stages of the RAG framework to enhance retrieval performance. For instance, SPLADE~\cite{formal2021splade} expands query with semantically similar terms, Self-RAG~\cite{asai2023self} performs self-correction on retrievals, 
RAT~\cite{wang2024rat} combines RAG with chain-of-thought reasoning, and LoRAG~\cite{thakur2024loops} improves text generation quality via iterative looping. Toolformer~\cite{schick2024toolformer} enables LLMs to call different tool APIs, allowing information gathering from diverse sources.

In the multi-modal domain, integrating RAG with LLMs remains relatively underexplored. FairRAG~\cite{shrestha2024fairrag} uses RAG to promote fairness and diversity in image generation, and RAR~\cite{liu2024rar} leverages RAG to assist in image classification and object detection. In the video domain, to our knowledge, only iRAG~\cite{arefeen2024irag} uses RAG by encoding video information into contextual natural language descriptions, enabling LLMs to interpret video content.

These observations indicate that RAG's application in the video domain remains very limited. RAG-Adapter is the first to directly integrate RAG with MLLMs, enhancing long video understanding at the input frame level.

\section{Method}

% \subsection{RAG Preliminaries}
% RAG is a prominent NLP approach that enhances language models' response generation by incorporating external information via a retrieval mechanism, which comprises two main stages: Data Preparation and Application.
% \vspace{-10pt}

% \paragraph{Data Preparation Stage.}
% In the case of text, the stage begins with downloading data from various sources, followed by filtering, compression, and other preprocessing steps. The text is then semantically segmented and encoded into embeddings, which are stored in a database.
% \vspace{-10pt}

% \paragraph{Application Stage.}
% The user provides a task-based prompt, and using similarity retrieval, the most relevant context is matched from the database and integrated with the prompt. The LLM then combines this prompt with the retrieved knowledge to generate a response.

\subsection{RAG-Adapter Pipeline}
\label{sec:pipeline}

RAG-Adapter is a simple yet effective plugin to enhance MLLMs' video understanding, with its main pipeline detailed in~\Cref{pipeline}. 
\vspace{-10pt}

\paragraph{Video Preprocessing.} 
For the test videos, frames are sampled at one frame per second, forming \(\{f_i\}_{i=1}^{N}\). Each frame is then encoded into image embeddings \(\{zf_i\}_{i=1}^{N}\) using the image encoder CLIP-L/14~\cite{radford2021learning}. As CLIP-L/14 primarily captures global features, which may miss fine-grained details like objects and actions, we also employ the open-source model CogVLM2~\cite{hong2024cogvlm2} to generate captions for each frame, resulting in the set \(\{c_i\}_{i=1}^{N}\). These captions are encoded into text embeddings \(\{zc_i\}_{i=1}^{N}\) using the text encoder BGE-M3~\cite{chen2024bge}, accommodating CLIP’s text length limitations. Here, \({f_i}\), \({zf_i}\), \({c_i}\), and \({zc_i}\) represent the \(i_{th}\) frame, its embedding, the caption, and its embedding, respectively. Finally, \(\{zf_i\}_{i=1}^{N}\) and \(\{zc_i\}_{i=1}^{N}\) are stored in the FramesDB and CaptionsDB databases for retrieval.
\vspace{-10pt}

\paragraph{Video Frames Retrieval.} 
To address the dimensional discrepancy between the text and image encoder embeddings and avoid the added complexity and potential performance issues of aligning these spaces, we employ a separate retrieval strategy. When a user submits a question, we encode it using both the text and image encoders and independently match it against the FramesDB and CaptionsDB, retrieving the Top\(M\) video frames \(\{f_i, sf_i\}_{i=1}^{M}\) and Top\(N\) captions \(\{c_i, sc_i\}_{i=1}^{N}\) from each respective databases, where \(sf_i\) and \(sc_i\) represent the similarity scores of the query with each retrieced frame and caption, respectively.

To effectively integrate the retrieval results from both databases, we introduce the {\bf Dual Reranker} module, comprising two main steps:

1) We sum the similarity scores of the Top\(M\) frames and Top\(N\) captions (noting that some captions may correspond to frames outside the Top\(M\) set), ranking them by these summed scores to obtain the Top\(X\) frames, their corresponding captions and scores, where X is determined jointly by M and N. The set \(\{f^{X}_{i}, c^{X}_{i}, s^{X}_{i}\}_{i=1}^{X}\) represents the \(i_{th}\) frame, its caption and summed score, respectively.

2) We find that frames ranked closely within the Top\(X\) often exhibit high similarity, reducing diversity. To maintain relevance while enhancing diversity, we apply the Maximal Marginal Relevance (MMR) algorithm~\cite{carbonell1998use}, commonly used in recommendation systems. We begin with an initially selected set \(\mathcal{S}=\emptyset\) and an unselected set  \(\mathcal{U}=\{f^{X}_{i}, c^{X}_{i}, s^{X}_{i}\}_{i=1}^{X}\). First, we add the frame with the highest summed score from \(\mathcal{U}\) to \(\mathcal{S}\). For each remaining frame in \(\mathcal{U}\), the one with the highest Marginal Relevance (MR) score, \(i^{\star}=\arg \max_{i \in \mathcal{U}} MR_{i}\), is then moved to \(\mathcal{S}\). This step is repeated \(K - 1\) times, producing \(K\) frames in \(\mathcal{S}\), representing Top\(K\) relevant frames selected by RAG-Adapter. The \(MR_{i}\) formula is as follow:
\vspace{-10pt}

\begin{equation}
    MR_{i} = \theta \cdot s^{X}_{i} - (1 - \theta) \cdot \mathop{max}\limits_{j \in \mathcal{S}} [ sim(f^{X}_i, f^{X}_j) + sim(c^{X}_i, c^{X}_j) ]
\end{equation}

\(\theta\) is a penalty coefficient to balance the weights of the summed similarity score and diversity score, with \(sim()\) computed via cosine similarity.

\begin{figure}[t]
    \vskip -0.1in
    \begin{center}
    \centerline{\includegraphics[width=0.8\linewidth]{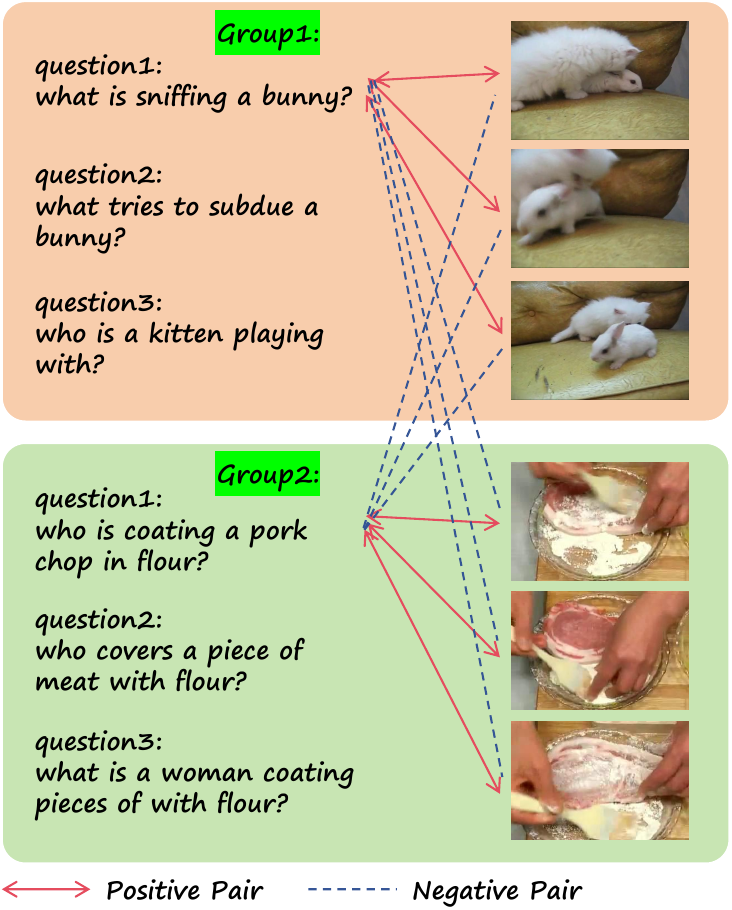}}
         \caption{Illustration of Grouped-supervised Contrastive Learning (GCL) constructing positive and negative pairs.}
     \label{loss}
     \end{center}
     \vskip -0.5in
\end{figure}

\subsection{RAG-Adapter Fine-tuning}
\label{sec:ft}

The text and image encoders in RAG-Adapter, BGE-M3 and CLIP-L/14, are trained on large-scale internet-based corpora. However, their embedding spaces may not be fully optimized for video understanding scenarios. To enhance RAG-Adapter's performance in this domain, we construct a specialized dataset, {\bf MMAT}, consisting of \((Q_i, F_i)\) and \((Q_i, C_i)\) positive pairs for contrastive learning fine-tuning of CLIP-L/14 and BGE-M3, respectively. Here, \(Q_i\), \(F_i\), and \(C_i\) denote the question, representative video frame, and corresponding caption for the \(i_{th}\) video.
\vspace{-10pt}

\paragraph{MMAT Construction.}
We employ a contrastive learning-based fine-tuning method to better fit BGE-M3 and CLIP-L/14's embedding spaces to the requirements of video understanding benchmarks. Given the challenge of identifying relevant frames in long videos, we start with widely used short video understanding benchmarks, including MSVD-QA~\cite{xu2017video}, MSRVTT-QA~\cite{xu2017video}, ActivityNet-QA~\cite{yu2019activitynet}, and TGIF-QA~\cite{jang2017tgif}, to construct the MMAT. To fully use the available videos, the training and validation sets from these benchmarks are combined to form the MMAT training set, while their test sets create the MMAT test set.

Since the videos in these benchmarks are typically short (usually under 10 seconds) with relatively consistent visual content, we sample frames at one frame per second and select three representative frames from quartile positions within each video. For each question related to the video, one of these frames is randomly chosen to construct the \((Q_i, F_i)\) pairs. For each \(F_i\), we use CogVLM2 to generate captions with detailed descriptions, thereby forming the corresponding \((Q_i, C_i)\) pairs. 

To ensure sampled frames align with questions despite potential content inconsistencies, we use a script to automatically exclude videos over 300 seconds and manually filter out those with visibly inconsistent visuals.

We also observe occasional garbled text from CogVLM2 when generating captions for frames with repetitive characters. To address this, we use a script to detect and either regenerate or manually correct such captions, followed by a quick review to ensure semantic consistency with the video frames. These measures ensure the quality of MMAT, resulting in \(\mathbf{417,993}\) \((Q_i, F_i)\) and \((Q_i, C_i)\) pairs in the training set and \(\mathbf{109,799}\) pairs in the test set.
\vspace{-10pt}

\paragraph{Grouped-supervised Contrastive Learning (GCL).}
In contrastive learning, a self-supervised loss is typically used, where only pairs like \((Q_i, F_i)\) and \((Q_i, C_i)\) are treated as positive samples, while all pairs \((Q_i, F_j)\) and \((Q_i, C_j)\) \(\{i \neq j\}\) are treated as negative samples by default. 

However, in video understanding scenarios, a single video may correspond to multiple questions. In Self-supervised Contrastive Learning, pairs like \((Q_i, F_j)\), \((Q_i, C_j)\) \(\{i \neq j\}\), which should be positive, may instead be treated as negative samples, disrupting training. As for Fully-supervised Contrastive Learning, label information is incorporated, but it requires the manual construction of negative pairs. 

To address this, we propose the {\bf Grouped-supervised Contrastive Learning (GCL)}, designed specifically for video understanding scenarios. In GCL, pairs from the same video are assigned a common group label. Within each group ``G'', all possible pairs, such as \((Q^{G}_i, F^{G}_j)\) and \((Q^{G}_i, C^{G}_k)\), are treated as positive, while pairs from different groups ``G'' and ``\(G'\)'', such as \((Q^{G}_i, F^{G'}_j)\) and \((Q^{G}_i, C^{G'}_k)\), are treated as negative. GCL iterates through all combinations, computes loss values for each, and then averages them. ~\Cref{loss} presents an illustration of GCL, with the loss function shown as follows:
\vspace{-15pt}

\begin{equation}
    \mathcal{L}_i = - \frac{1}{|P(i)|} \sum_{p \in P(i)} \log \frac{\exp(\text{sim}(Q_i, {F_p/C_p}) / \tau)}{\sum_{a \in A'(i)} \exp(\text{sim}(Q_i, {F_a/C_a}) / \tau)}
\end{equation}

\begin{equation}
    \mathcal{L}_{GCL} = \frac{1}{|I|} \sum_{i \in I} \mathcal{L}_i
\end{equation}

Here, \(I\) represents the set of all test questions, \(\mathcal{L}_i\) denotes the loss function for \(Q_i\), and \(P(i)\) is the set of positive samples associated with \(Q_i\) within the same group. The notation \(A'(i) = A(i) \setminus \{ p' \in P(i) \setminus \{ p \} \}\) refers to the set of all batch samples (\(A(i)\)), excluding all positive samples except the selected one, \(F_p\) or \(Q_p\). \(\tau\) is the temperature coefficient. 
% The similarity function \(sim()\) is computed in parallel via dot product to enhance processing speed.

This approach offers two main advantages: (1) it eliminates the need for manually constructing numerous negative pairs, and (2) by excludes all positive samples but the selected one \(F_p\) or \(C_p\) from \(L_i\)'s denominator, the model better captures detailed relationships between the question and each specific positive sample, facilitating a more refined understanding of each sample's unique characteristics.
\section{Experiments and Analysis}

\begin{table}[t]
% \vskip -0.15in
\caption{ASS and NIF metrics for evaluation benchmarks.}
    \vspace{-15pt}
    \label{nie}
    % \vskip -0.5in
    \begin{center}
    % \begin{small}
    % \begin{sc}
\resizebox{0.9\linewidth}{!}
{
\begin{tabular}{c|c|c|c|c}
\hline
\multirow{2}{*}{Benchmarks} & \multicolumn{3}{c|}{ASS (0-2)} & \multirow{2}{*}{NIF} \\  \cline{2-4}
 & Top10 & Top30 & Top50 \\ \hline
 Video-MME & 0.81 & 0.69 & 0.61 & 2.4 \\ \hline
 MLVU & 0.76 & 0.61 & 0.52 & 2.9 \\ \hline
 Perception Test & 0.74 & 0.69 & 0.68 & 1.9 \\ \hline
 EgoSchema & 0.82 & 0.67 & 0.58 & 2.1 \\ \hline
\end{tabular}
}
    % \end{sc}
    % \end{small}
    \end{center}
    \vskip -0.3in
\end{table}

\begin{table*}[t]
% \vskip -0.15in
\caption{The test results for various MLLMs on Video-MME include accuracy metrics for 6 domains, along with the overall \textbf{Avg. Acc.} (Average Accuracy). The highest accuracy is \textbf{bolded}, and the second highest is \underline{underlined}.}
    \vspace{-15pt}
    \label{experiment1}
    % \vskip -0.15in
    \begin{center}
    % \begin{small}
    % \begin{sc}
\resizebox{0.8\linewidth}{!}{
\begin{tabular}{c|c|c|c|c|c|c|c|c}
\hline \hline
\multirow{2}{*}{Models}& \multirow{2}{*}{Sampling Method} & \multicolumn{6}{c|}{Category} & \multirow{2}{*}{Avg. Acc. (\%)}\\ \cline{3-8}
&  & Knowledge & Film \& Television & Sports Competition & Artistic Performance & Life Record & Multilingual & \\ \hline
\multicolumn{9}{c}{\textcolor[RGB]{128,128,128}{\textit{Image MLLMs}}} \\ \hline

\multirow{2}{*}{Otter-I~\cite{li2023otter}} & Uniform  & 28.9 & 28.9 & 33.3 & 33.3 & 24.4 & 33.3 & 30.4  \\
 & RAG-Adapter  & 37.8 (\textit{+8.9})& 37.8 (\textit{+8.9})& 42.2 (\textit{+8.9})& 40.0 (\textit{+6.7})& 28.9 (\textit{+4.5})& 31.1 (\textit{-2.2})& 36.3 (\textit{+5.9}) \\ \hline
 
\multirow{2}{*}{LLaVA-1.6~\cite{liu2024llavanext}} &Uniform & 33.3 & 22.2 & 33.3 & 44.4 & 26.7 & 24.4 & 30.7  \\
 & RAG-Adapter& 35.6 (\textit{+2.3})& 28.9 (\textit{+6.7})& 37.8 (\textit{+4.5})& 48.9 (\textit{+4.5})& 31.1 (\textit{+4.4})& 31.1 (\textit{+6.7})& 35.6 (\textit{+4.9})\\ \hline

\multirow{2}{*}{GPT4-Turbo~\cite{achiam2023gpt}} &Uniform &  60.0 & \underline{71.1} &48.9 & 57.8 & 53.3 & \underline{55.6}  & 57.8  \\ 
 &RAG-Adapter & \underline{71.1} (\textit{+11.1})& \underline{71.1} (\textit{+0.0})& 51.1 (\textit{+2.2})& 60.0 (\textit{+2.2})& 53.3 (\textit{+0.0})& \textbf{60.0} (\textit{+4.4})& 61.1 (\textit{+3.3}) \\ \hline \hline
 
\multicolumn{9}{c}{\textcolor[RGB]{128,128,128}{\textit{Video MLLMs}}} \\ \hline

\multirow{2}{*}{Otter-V~\cite{li2023otter}} & Uniform  & 31.1 & 28.9 & 33.3 & 22.2 & 31.1 & 26.7 & 28.9  \\
 & RAG-Adapter  & 37.8 (\textit{+6.7}) & 31.1 (\textit{+2.2})& 35.6 (\textit{+2.3})& 28.9 (\textit{+6.7})& 37.8 (\textit{+6.7})& 31.1 (\textit{+4.4})& 33.7 (\textit{+4.8}) \\ \hline

\multirow{2}{*}{mPlug-Owl-V~\cite{ye2023mplug}} & Uniform & 22.2 & 31.1 & 24.4 & 28.9 & 17.8 & 24.4 & 24.8  \\
 & RAG-Adapter & 35.6 (\textit{+13.4})& 37.8 (\textit{+6.7})& 28.9 (\textit{+4.5})& 31.1 (\textit{+2.2})& 26.7 (\textit{+8.9})& 28.9 (\textit{+4.5})& 31.5  (\textit{+6.7})\\ \hline

\multirow{2}{*}{MovieChat~\cite{song2024moviechat}} &Uniform  & 24.4 & 28.9 & 22.2 & 31.1 & 22.2 & 28.9 & 26.3  \\
 & RAG-Adapter & 33.3 (\textit{+8.9})& 35.6 (\textit{+6.7})& 33.3 (\textit{+11.1})& 31.1 (\textit{+0.0})& 28.9 (\textit{+6.7})& 35.6 (\textit{+6.7})& 33.0 (\textit{+6.7}) \\ \hline

\multirow{2}{*}{VideoChat~\cite{li2023videochat}} &Uniform  & 26.7 & 28.9 & 33.3 & 26.7 & 37.8 & 22.2 & 29.3  \\
 & RAG-Adapter & 31.1 (\textit{+4.4})& 35.6 (\textit{+6.7})& 33.3 (\textit{+0.0})& 33.3 (\textit{+6.6})& 40.0 (\textit{+2.2})& 33.3 (\textit{+11.1})& 34.4 (\textit{+5.1}) \\ \hline

\multirow{2}{*}{VideoChat2~\cite{li2024mvbench}} & Uniform & 33.3 & 17.8 & 24.4 & 35.6 & 44.4 & 28.9 & 30.7 \\
 & RAG-Adapter & 42.2 (\textit{+8.9})& 22.2 (\textit{+4.4}) & 26.7 (\textit{+2.3})& 37.8 (\textit{+2.2})& 42.2 (\textit{-2.2})& 31.1 (\textit{+2.2})& 33.7 (\textit{+3.0})\\ \hline

\multirow{2}{*}{LLaMA-VID~\cite{li2025llama}} & Uniform & 31.1 & 17.8 & 24.4 & 37.8 & 22.2 & 26.7 & 26.7  \\
 & RAG-Adapter & 31.1 (\textit{+0.0}) & 26.7 (\textit{+8.9}) & 33.3 (\textit{+8.9})& 37.8 (\textit{+0.0})& 28.9 (\textit{+6.7})& 28.9 (\textit{+2.2})& 31.1 (\textit{+4.4}) \\ \hline

\multirow{2}{*}{TimeChat~\cite{ren2024timechat}} & Uniform & 31.1 & 33.3 & 28.9 & 46.7 & 31.1 & 26.7 & 33.0  \\
 & RAG-Adapter & 33.3  (\textit{+2.2})& 42.2  (\textit{+8.9})& 31.1  (\textit{+2.2})& 48.9  (\textit{+2.2})& 33.3  (\textit{+2.2})& 33.3  (\textit{+6.6})& 37.0  (\textit{+4.0}) \\ \hline

 \multirow{2}{*}{Chat-UniVi~\cite{jin2024chat}} & Uniform & 33.0 & 26.7 & 24.4 & 37.8 & 31.1 & 24.4 & 29.6  \\
 & RAG-Adapter & 42.2  (\textit{+8.9})& 35.6  (\textit{+8.9})& 35.6  (\textit{+11.2})& 46.7  (\textit{+8.9})& 42.2  (\textit{+11.1})& 35.6  (\textit{+11.2})& 39.7  (\textit{+10.1}) \\ \hline

\multirow{2}{*}{GPT-4o~\cite{openai}} & Uniform & 66.7 & 62.2 & \underline{66.7} & \underline{64.4} & \underline{55.6} & 53.5 & \underline{61.5} \\
 & RAG-Adapter & \textbf{77.8} (\textit{+11.1})& \textbf{73.3} (\textit{+11.1})& \textbf{68.9} (\textit{+2.2})& \textbf{75.6} (\textit{+11.2})& \textbf{68.9} (\textit{+13.3})& \textbf{60.0} (\textit{+6.7})& \textbf{70.8} (\textit{+9.3})\\ \hline \hline
\end{tabular}
}
    % \end{sc}
    % \end{small}
    \end{center}
    \vskip -0.3in
\end{table*}

\begin{table}[t]
% \vskip -0.15in
\caption{Comparison across more benchmarks.}
\vspace{-15pt}
    \label{other benchmark}
    % \vskip -0.15in
    \begin{center}
    % \begin{small}
    % \begin{sc}
\resizebox{0.95\linewidth}{!}{
\begin{tabular}{c|c|c|c|c}
\hline \hline
Models& Sampling Method & MLVU & Perception Test & EgoSchema \\ \hline

\multirow{2}{*}{MovieChat~\cite{song2024moviechat}} &Uniform  & 29.6 & 32.5 &  23.3  \\
 & RAG-Adapter & 41.5 (\textit{+11.9})& 37.8 (\textit{+5.3})& 28.9 (\textit{+5.6})\\ \hline

\multirow{2}{*}{LLaMA-VID~\cite{li2025llama}} & Uniform & 34.8 & 33.1 & 24.4  \\
 & RAG-Adapter & \underline{43.0} (\textit{+8.2}) & 37.2 (\textit{+4.1})& 31.1 (\textit{+6.7})\\ \hline

\multirow{2}{*}{TimeChat~\cite{ren2024timechat}} & Uniform & 37.8 & 37.8 & 27.8   \\
 & RAG-Adapter & \textbf{45.2} (\textit{+7.4})& \underline{41.1} (\textit{+3.3})&\underline{32.2} (\textit{+4.4})\\ \hline

\multirow{2}{*}{Chat-UniVi~\cite{jin2024chat}} & Uniform & 32.6 & 38.1 & \underline{32.2} \\
 & RAG-Adapter & 40.0 (\textit{+7.4})&  \textbf{41.6} (\textit{+3.5})& \textbf{41.1} (\textit{+8.9})\\ \hline \hline
\end{tabular}
}
    % \end{sc}
    % \end{small}
    \end{center}
    \vskip -0.3in
\end{table}

\subsection{Evaluation Benchmarks}
\label{evaluation benchmarks}

We select two commonly used long video understanding benchmarks, Video-MME and MLVU, as well as two relatively shorter benchmarks focusing on human interaction and perception in real-world scenarios: Perception Test~\cite{patraucean2023perception} and EgoSchema~\cite{mangalam2023egoschema}, for evaluation. This selection aims to demonstrate RAG-Adapter's generalization performance across benchmarks with varying temporal spans (ranging from approximately 0.5 minutes to several hours) and contexts. 
Due to the time-consuming process of generating captions for each video (as discussed in~\Cref{component}), we sample 90 videos from each benchmark to manage this.
% As the RAG-Adapter pipeline samples one frame per second and generates corresponding captions, annotating hour-long videos would involve numerous frames, with an average 
% offline processing time of approximately 1-2 hours (More discussions can be seen in the supplementary materials). To manage this, we sample 90 videos from each benchmark. 
Video-MME videos are categorized by length (short, medium, long) and further divided into six content domains, from which we randomly select five videos per domain. MLVU videos are classified into nine types, from which ten videos are randomly sampled per category. For Perception Test, the 90 longest videos from the Multiple-Choice Video Q\&A task are selected. In Egoschema, where all videos are 3 minutes long, 90 videos are randomly sampled.

\subsection{Statistics of ASS and NIF}
Using the RAG-Adapter, we calculate the ASS and NIF metrics for all evaluation benchmarks.

For ASS, we measure the average summed score, \(ASS=\frac{1}{n}\sum_{i=1}^{K}s^{i}\) (on a 0-2 scale), between all questions and their Top\(K\) relevant frames with captions, where \(K\) set to 10, 30, and 50. For NIF, we manually identify the minimum number of frames containing essential visual information needed to answer each question (note: for Video-MME, some information is also found in subtitle files). The NIF value is the average of these frame counts across all questions. Results are summarized in~\Cref{nie}. 
% first retrieve the Top\(50\) relevant frames, then

The ASS values for the four benchmarks are similar, as the top\(K\) frames retrieved by the RAG-Adapter in each benchmark show little variation in relevance to the questions. Additionally, as K increases, the overall relevance tends to decrease. MLVU has the highest NIF value due to the greater frame requirement for Action Order and Video Summarization tasks. Both MLVU and Video-MME show slightly higher NIF values than the other two benchmarks, given their longer average durations, though the number of frames containing essential information remains limited. Supplementary materials provide each test video's ID, corresponding question (or ID), minimum frame count, frame timestamps and identified issues in the benchmarks using RAG-Adapter.

% The NIF value for MLVU is higher than for Video-MME, as MLVU contains questions related to Action Order and Video Summarization require more key frames. Action Order questions typically require around 6 frames, while Video Summarization often needs 10-20 frames. In contrast, Video-MME shows a slightly higher ASS value, suggesting its questions are more directly related to the video content. Supplementary materials provide each test video's ID, corresponding question (or ID), minimum frame count, and frame timestamps for reference, along with several unreasonable issues identified in the test benchmarks using RAG-Adapter.

\subsection{Baselines and Experimental Setups}
We classify the baselines into two categories: image-based MLLMs supporting multiple image inputs and video MLLMs. Open-source models are tested locally on an NVIDIA 4090 GPU, while proprietary models are accessed via official APIs. 
Based on the NIF metrics (\Cref{nie}) and the maximum number of key frames—Video-MME (9), MLVU (20), Perception Test (8), and EgoSchema (6)—we set \(K=10\) frames for Video-MME, Perception Test, and EgoSchema, and \(K=20\) for MLVU to ensure sufficient information retrieval by RAG-Adapter and maintain evaluation fairness (\(M\), \(N\), and \(\theta\) are set to \(50\), \(50\), and \(0.7\), respectively). Frames are input in chronological order to preserve temporal information. 
We compare each MLLM's performance under identical frame input conditions, contrasting uniform sampling with RAG-Adapter sampling.
The accuracy (ranging from 0 to 100) for multiple-choice questions across the four benchmarks is calculated by comparing the predicted results with the ground truth.
% For multiple-choice questions, accuracy (0–100) is calculated by matching predicted results with the ground truth. 
% For open-ended questions (present in MLVU), we apply the original benchmark's evaluation criteria and use GPT-4o~\cite{openai} to rank generated answers on a scale of 0 to 10.

\begin{table}[t]
% \vskip -0.15in
\caption{Comparison under different fine-tuning methods.}
\vspace{-15pt}
    \label{experiment2}
    % \vskip -0.15in
    \begin{center}
    % \begin{small}
    % \begin{sc}
\resizebox{0.75\linewidth}{!}{
\begin{tabular}{c|c|c}
\hline \hline
Models& Fine-Tuning Method &  Avg. Acc. (\%)  \\ \hline

% \multirow{4}{*}{MovieChat~\cite{song2024moviechat}} & No Fine-Tuning  & 29.2 \\ \cline{2-3}
%  & SCL & 28.5 (\textit{-0.7})  \\ \cline{2-3}
%  & CB & 29.3 (\textit{+0.1})\\ \cline{2-3}
%  & GCL &  \textbf{33.0} (\textit{+3.8}) \\ \hline

%  \multirow{4}{*}{LLaMA-VID~\cite{li2025llama}} & No Fine-Tuning  & 28.9  \\ \cline{2-3}
%  & SCL &  28.9 (\textit{+0.0}) \\ \cline{2-3}
%  & CB & 29.6 (\textit{+0.7})\\ \cline{2-3}
%  & GCL & \textbf{31.1} (\textit{+2.2}) \\ \hline

 \multirow{4}{*}{TimeChat~\cite{ren2024timechat}} & No Fine-Tuning  & 33.7 \\ \cline{2-3}
 & SCL & 32.6 (\textit{-1.1}) \\ \cline{2-3}
  &CB & 34.8 (\textit{+1.1})\\ \cline{2-3}
 & GCL & \textbf{37.0} (\textit{+3.3}) \\ \hline

 \multirow{4}{*}{GPT-4o~\cite{openai}} & No Fine-Tuning  & 65.9 \\ \cline{2-3}
 & SCL & 65.2 (\textit{-0.7}) \\ \cline{2-3}
  & CB & 66.7 (\textit{+0.8})\\ \cline{2-3}
 & GCL & \textbf{70.8} (\textit{+4.9}) \\ \hline \hline

\end{tabular}
}
    % \end{sc}
    % \end{small}
    \end{center}
    \vskip -0.4in
\end{table}

\begin{table*}[t]
% \vskip -0.15in
\caption{Ablation Study of the RAG-Adapter Components, where ``T\&E'' denotes Text Encoder, ``I\&E'' represents Image Encoder, and ``D\&R'' stands for the Dual Ranker.}
\vspace{-15pt}
    \label{component}
    % \vskip -0.15in
    \begin{center}
    % \begin{small}
    % \begin{sc}
\resizebox{0.7\linewidth}{!}{
\begin{tabular}{c|c|c|c|c|c|c|c|c}
\hline \hline
\multicolumn{9}{c}{\textcolor[RGB]{128,128,128}{\textit{Component Ablation}}} \\ \hline

Fine-Tuning Method & T\&E & I\&E &  D\&R  & Avg. Acc. (\%) & Preprocessing & Retrieval & Inference & Recall@10 \\ \hline

\multirow{4}{*}{GCL} & \checkmark  &  & & 33.3 & 48.8min & 4.7s &\multirow{7}{*}{0.88s} & 19.7 \\ \cline{5-7}  \cline{9-9}
 &  &\checkmark & & 34.5 &11.4s & 8.7s & & 18.7 \\ \cline{5-7}  \cline{9-9}
  & \checkmark & \checkmark & & 35.2 & \multirow{5}{*}{48.8min} & 13.5s & & 24.8 \\ \cline{5-5} \cline{7-7} \cline{9-9}
  & \checkmark & \checkmark &\checkmark  & \textbf{39.7} & & \multirow{4}{*}{15.4s} &  & \textbf{30.4} \\ \cline{1-5} \cline{9-9}
 No Fine-Tuning & \checkmark &\checkmark &\checkmark  & 33.3 & & &  & 24.1 \\ \cline{1-5} \cline{9-9}
 SCL & \checkmark &\checkmark &\checkmark  & 32.2 &  &  & & 23.7 \\  \cline{1-5} \cline{9-9}
 CB & \checkmark &\checkmark &\checkmark  & 35.6 & &  & & 25.1 \\
 \hline \hline
 \multicolumn{9}{c}{\textcolor[RGB]{128,128,128}{\textit{Other Baselines}}} \\ \hline
  \multicolumn{4}{c|}{Sampling Method} & Avg. Acc. (\%) & Preprocessing & Retrieval & Inference & Recall@10 \\ \hline
  \multicolumn{4}{c|}{Uniform} & 29.6 & 11.4s & N/A &\multirow{2}{*}{0.88s} & 5.5 \\ \cline{1-7} \cline{9-9}
  % \multicolumn{4}{c|}{Segment by MLLM} & & 11.4s & 98.2s & & \\ \cline{1-7} \cline{9-9}
  \multicolumn{4}{c|}{Two-stage Retrieval} & \underline{37.0} & 4.3min & (8.5+2.5)s & & \underline{25.8} \\ \hline \hline
 
\end{tabular}
}
    % \end{sc}
    % \end{small}
    \end{center}
    \vskip -0.3in
\end{table*}

% (numbers in parentheses are relative to uniform sampling with 10 frames for each model)
\begin{table}[t]
% \vskip -0.15in
\caption{Comparison of different sampling strategies and input frame counts.}
\vspace{-15pt}
    \label{experiment3}
    % \vskip -0.15in
    \begin{center}
    % \begin{small}
    % \begin{sc}
\resizebox{0.90\linewidth}{!}{
\begin{tabular}{c|c|c|c}
\hline \hline
Models& Sampling Method & Frames Count & Avg. Acc. (\%)\\ \hline

% \multirow{7}{*}{MovieChat~\cite{song2024moviechat}} 
% & \multirow{4}{*}{Uniform} & 5 & 25.6  \\ \cline{3-4}
% &  & 10 & 26.3  \\ \cline{3-4}
% &  & 20 & 27.0 \\ \cline{3-4}
% &  & 512 & 28.9 (\textit{+2.6}) \\ \cline{2-4}
% & \multirow{3}{*}{RAG-Adapter} & 5 & 30.0 (\textit{+3.7})\\ \cline{3-4}
% &  & 10 & 33.0 (\textit{+6.7})\\ \cline{3-4}
%  &  & 20 & \textbf{33.3} (\textit{+7.0})\\ \hline

%  \multirow{7}{*}{LLaMA-VID~\cite{li2025llama}}
%  & \multirow{4}{*}{Uniform} & 5 & 26.7  \\ \cline{3-4}
%  &  & 10 & 26.7  \\ \cline{3-4}
%  &  & 20 & 27.1  \\ \cline{3-4}
% &  & 512 & 27.8 (\textit{+1.1})\\ \cline{2-4}
% & \multirow{3}{*}{RAG-Adapter} & 5 & 28.9 (\textit{+2.2})\\ \cline{3-4}
% &  & 10 & \textbf{31.1} (\textit{+4.4})\\ \cline{3-4}
%  &  & 20 & 30.8 (\textit{+4.1})\\ \hline

 \multirow{7}{*}{TimeChat~\cite{ren2024timechat}}
 & \multirow{4}{*}{Uniform} & 5 & 30.7  \\ \cline{3-4}
 &  & 10 & 33.0  \\ \cline{3-4}
 &  & 20 & 33.0  \\ \cline{3-4}
&  & 64 & 32.2 \\ \cline{2-4}
& \multirow{3}{*}{RAG-Adapter} & 5 & 36.3 \\ \cline{3-4}
&  & 10 & \textbf{37.0} \\ \cline{3-4}
 &  & 20 & \textbf{37.0} \\ \hline

\multirow{7}{*}{Chat-UniVi~\cite{jin2024chat}}
&  \multirow{4}{*}{Uniform} & 5 & 27.8  \\ \cline{3-4}
&  & 10 & 29.6  \\ \cline{3-4}
& & 20 & 32.6  \\ \cline{3-4}
&  & 256 & 31.9  \\ \cline{2-4}
& \multirow{3}{*}{RAG-Adapter} & 5 & 35.6 \\ \cline{3-4}
&  & 10 & 39.7 \\ \cline{3-4}
 &  & 20 & \textbf{40.0} \\ \hline \hline

\end{tabular}
}
    % \end{sc}
    % \end{small}
    \end{center}
    \vskip -0.3in
\end{table}

\begin{table}[t]
% \vskip -0.15in
\caption{Comparison between no subtitles (w/o subs), subtitles corresponding to RAG-Adapter sampled frames (w/ subs (Corresp.)), and subtitles sampled by RAG-Adapter (w/ subs (RAG-Adapter)).}
\vspace{-15pt}
    \label{experiment4}
    % \vskip -0.15in
    \begin{center}
    % \begin{small}
    % \begin{sc}
\resizebox{0.9\linewidth}{!}{
\begin{tabular}{c|c|c}
\hline \hline
Models& Subtitles &  Avg. Acc. (\%)\\ \hline

% \multirow{3}{*}{MovieChat}~\cite{song2024moviechat} & w/o subs & 33.0  \\ \cline{2-3}
%  & w/ subs (Corresp.) &  34.1 (\textit{+1.1}) \\ \cline{2-3}
%  & w/ subs (RAG-Adapter) &  \textbf{34.8} (\textit{+1.8}) \\ \hline

%  \multirow{3}{*}{LLaMA-VID}~\cite{li2025llama} & w/o subs & 31.1 \\ \cline{2-3}
%  & w/ subs (Corresp.) &  31.9 (\textit{+0.8})\\ \cline{2-3}
%  & w/ subs (RAG-Adapter) & \textbf{32.2} (\textit{+1.1})\\ \hline

 \multirow{3}{*}{TimeChat~\cite{ren2024timechat}} & w/o subs  & 37.0 \\ \cline{2-3}
 & w/ subs (Corresp.) & 38.2 (\textit{+1.2})\\ \cline{2-3}
 & w/ subs (RAG-Adapter) & \textbf{39.6} (\textit{+2.6})\\ \hline

 \multirow{3}{*}{Chat-UniVi~\cite{jin2024chat}} & w/o subs & 39.7 \\ \cline{2-3}
 & w/ subs (Corresp.) & 40.0 (\textit{+0.3})\\ \cline{2-3}
 & w/ subs (RAG-Adapter) & \textbf{41.5} (\textit{+1.8})\\ \hline \hline

\end{tabular}
}
    % \end{sc}
    % \end{small}
    \end{center}
    \vskip -0.3in
\end{table}

\subsection{Results and Analysis}
\label{results and analysis}
\Cref{experiment1} compares the performance of uniform sampling and RAG-Adapter sampling across six domains in the Video-MME (without subtitle information). 
\Cref{other benchmark} compares the performance on the remaining three benchmarks, and the more comprehensive experimental results for MLVU are provided in the supplementary materials.
% Test results for MLVU are available in the supplementary materials. 
From the experimental results, we draw the following key conclusions:

\begin{itemize}
\item \textbf{Performance:} RAG-Adapter sampling improves overall performance across all models compared to uniform sampling, indicating that the information loss from uniform sampling adversely affects model performance in testing. Thus, uniform sampling does not fully reflect MLLMs' true long video understanding capabilities.

% \item The accuracy of open-source image-based MLLMs and video-based MLLMs is comparable, suggesting that video-based MLLMs do not offer a significant advantage when the number of input frames is limited. Our ablation studies (\Cref{experiment3}) further examine how varying input frame counts affect the accuracy of video-based MLLMs.

\item \textbf{Unified Improvement:} In~\Cref{experiment1}, while commercial MLLMs outperform open-source models, RAG-Adapter consistently enhances performance. For example, GPT-4o shows accuracy gains exceeding 10\% in the Knowledge, Film \& Television, Artistic Performance, and Life Record domains, with an average accuracy increase of 9.3\%. Models like mPLUG-Owl-V show a 13.4\% improvement in Knowledge, with an overall accuracy increase of 6.7\%, while Chat-UniVi achieves over a 10\% improvement in Sports Competition, Life Record, Multilingual, and overall accuracy increase. This demonstrates the versatility of RAG-Adapter, as its effectiveness is not directly linked to the intrinsic capabilities of the models.

\item \textbf{Generalization:} In~\Cref{other benchmark}, Perception Test and Egochema have shorter average durations (35s\&3min) compared to MLVU (12min), leading to a less pronounced improvement of RAG-Adapter over uniform sampling. Nonetheless, its performance across benchmarks of varying lengths demonstrates the method’s effectiveness and generalization.

\item \textbf{Constraint:} In~\Cref{experiment1}, RAG-Adapter does not consistently improve accuracy across all domains. GPT-4 Turbo shows no improvement in Film \& Television, while MovieChat and LLaMA-VID remain unchanged in Artistic Performance. Otter-I and VideoChat2 experience a 2.2\% accuracy drop in Multilingual and Life Record. This stability or slight decline is primarily due to RAG-Adapter occasionally failing to retrieve all relevant information, resulting in the omission of key frames. In such cases, RAG-Adapter may mislead the model, affecting its accuracy. We aim to further refine the retrieval process of RAG-Adapter to minimize these issues.

\end{itemize}
% \vspace{-5pt}

\begin{figure*}[t]
    % \vskip -0.1in
    \begin{center}
    \centerline{\includegraphics[width=0.9\linewidth]{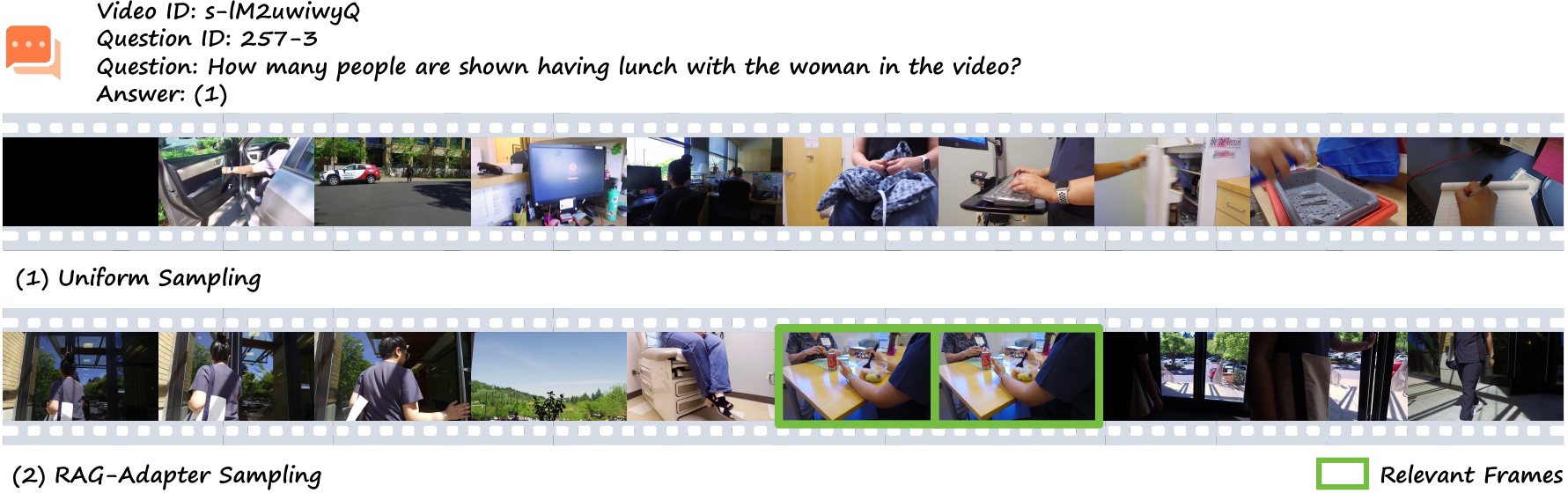}}
     \caption{Comparison of RAG-Adapter and uniform sampling results: RAG-Adapter accurately identifies two consecutive key frames relevant to the question, whereas uniform sampling tends to miss them.}
     \label{visualization}
     \end{center}
     \vskip -0.4in
\end{figure*}

\begin{figure}[t]
    % \vskip -0.1in
    \begin{center}
    \centerline{\includegraphics[width=0.9\linewidth]{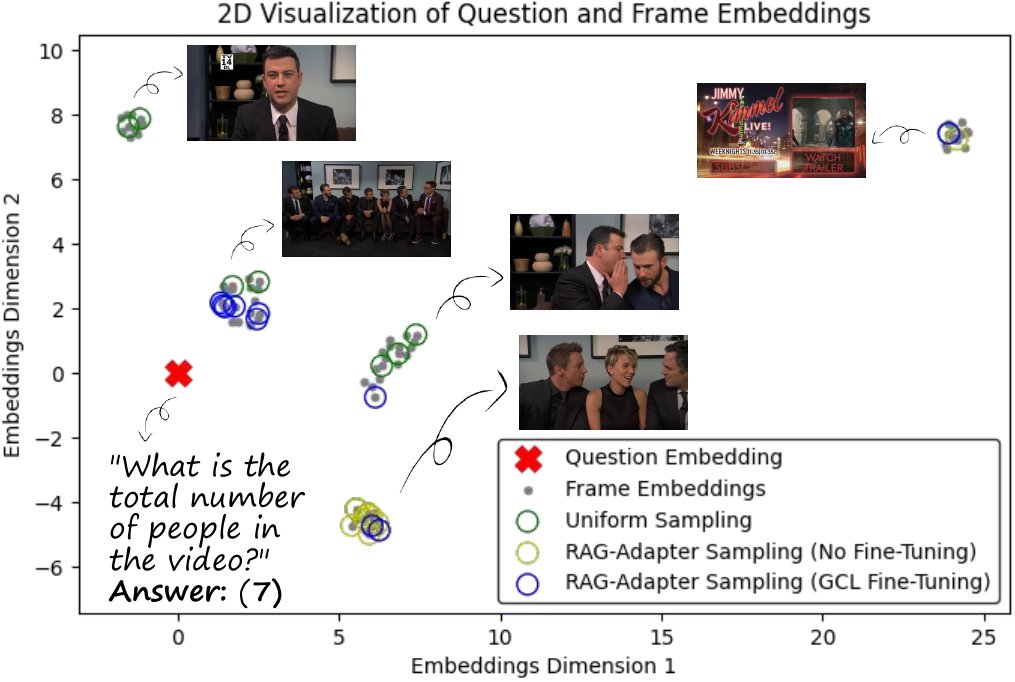}}
         \caption{The relationship between the embedding spaces of video frames sampled using different methods and that of the corresponding questions. The frame embeddings are primarily grouped into five clusters, each representing a set of consecutive shots, with each cluster labeled by a representative frame.}
     \label{embedding}
     \end{center}
     \vskip -0.4in
\end{figure}

\subsection{Ablation Study}
The following ablation experiments are conducted on the Video-MME benchmark, with additional results provided in the supplementary materials.
\vspace{-15pt}

\paragraph{Effect of RAG-Adapter Fine-Tuning.}
In~\Cref{experiment2}, we evaluate RAG-Adapter's performance across different fine-tuning approaches, including No Fine-Tuning, Self-supervised Contrastive Learning (SCL) fine-tuning, Customizing Batch (CB) fine-tuning (where each question in a batch belongs to a different video), and Grouped-supervised Contrastive Learning (GCL) fine-tuning. The results demonstrate that GCL fine-tuning achieves superior performance for all models. This enhancement is primarily due to GCL's ability to train RAG-Adapter's text and image encoders to effectively learn rich positive sample features within each group while avoiding the adverse effects of treating intra-group samples as negatives, as seen in SCL. Moreover, GCL retains all inter-group negative samples from SCL and CB, ensuring robust learning.
\vspace{-10pt}

\paragraph{Discussion on the effectiveness and efficiency of RAG-Adapter components.}
In~\Cref{component}, we utilize Chat-UniVi to conduct an ablation study on RAG-Adapter's components, evaluating pipeline efficiency (scaled to 10-minute videos per query) and average recall. The preprocessing phase involves frame sampling and captioning. Our method achieves optimal accuracy and recall only with all components and GCL fine-tuning. 
To address the impracticality of captioning every frame in long videos, we propose a \textbf{Two-stage Retrieval} method: initially retrieving the top\(50\) frames, then using captions to refine the selection to the final top\(K\) frames, achieving a favorable balance between accuracy and efficiency.
Also, retrieval accuracy using only Image Encoder outperforms uniform sampling, offering another viable alternative in practice.
Furthermore, inspired by SeViLA~\cite{yu2023self}, we employ the tested MLLM for frame filtering but find it time-intensive and ineffective.
% as it labels 90\% of frames as relevant, exceeding the MLLM's context length constraints.
\vspace{-10pt}

\paragraph{Comparison of Different Input Frame Counts.}
Since some long-video MLLMs can support a larger number of input frames, we compare uniform sampling (using 5, 10, 20 frames or the model's maximum supported frames on a single NVIDIA 4090) with RAG-Adapter sampling (using 5, 10, and 20 frames), as shown in~\Cref{experiment3}. Results indicate that, despite utilizing more frames, uniform sampling does not outperform RAG-Adapter and even exhibits slight performance degradation compared to using fewer frames. For RAG-Adapter sampling, performance improves from \(K=5\) to \(K=10\), suggesting information loss at \(K=5\), and stabilizes at \(K=20\). This aligns with the NIF metric for Video-MME, which is below 10, implying most questions require fewer than 10 frames to capture essential information. Additionally, increasing uniformly sampled frames does not guarantee inclusion of critical details and may introduce greater redundancy.
\vspace{-10pt}

\paragraph{Impact of Subtitle Information.}
In the Video-MME benchmark, subtitle files are available and contain some information relevant to certain questions. In~\Cref{experiment4}, we examine the impact of subtitles on MLLMs using 10 frames sampled by RAG-Adapter. We evaluate two subtitle inclusion methods: providing subtitles directly correspond to the sampled frames and using RAG-Adapter to select the 10 most relevant subtitles for each question.

Our experiments reveal two main insights. First, model accuracy consistently improves with subtitle inclusion, as subtitles often provide question-relevant information. Second, subtitles filtered by RAG-Adapter outperform those directly tied to sampled frames, as critical subtitle information may not align with key video content, and complex questions often rely more heavily on subtitle data.
% \vspace{-10pt}

\section{Visualization}

% In this section, we compare video frames obtained through uniform sampling with those sampled using the RAG-Adapter. Additionally, we analyze the embedding spaces differences between the uniformly sampled frames and RAG-Adapter sampled frames, both without and with GCL fine-tuning.

\subsection{Visualization of Frame Sampling Methods}
In~\Cref{visualization}, we compare the results of uniform sampling and RAG-Adapter sampling for the same question in the Video-MME benchmark. The specific scene referenced in the question - ``How many people are shown having lunch with the woman in the video?'', occurs only between 73-74 seconds in the original video. As a result, uniform sampling fails to capture any relevant frames, whereas RAG-Adapter successfully identifies the two pertinent frames (sampled at one frame per second). Additional visualizations of the video frames are provided in the supplementary materials.

\subsection{Differences of Embedding spaces}
In~\Cref{embedding}, we reduce the embedding space of the question and all video frames to two dimensions using UMAP~\cite{mcinnes2018umap} (Uniform Manifold Approximation and Projection) to preserve the global structure of the data. This visualization illustrates the spatial relationship between the question embedding and the embeddings of frames sampled by uniform sampling, the non-fine-tuned RAG-Adapter, and the GCL fine-tuned RAG-Adapter. It can be observed that the embeddings of uniformly sampled frames are highly scattered, while the embeddings of frames sampled by the non-fine-tuned RAG-Adapter cluster around a few similar frames. In contrast, the embeddings from the GCL fine-tuned RAG-Adapter exhibit greater diversity and are closer to the question embedding.
\section{Conclusion}

In this paper, we integrate the RAG framework with MLLMs, introducing RAG-Adapter, a plugin that enhances the long video understanding capabilities of MLLMs without modifying their internal structure. By providing question-relevant video frames during testing, RAG-Adapter ensures that the evaluation of long video understanding benchmarks accurately reflects the model’s true video comprehension capabilities. To better adapt RAG-Adapter to the video question-answering context, we construct a fine-tuning dataset, MMAT, and introduce Grouped-supervised Contrastive Learning (GCL) to help RAG-Adapter learn rich and relevant embedding between questions and video frames. Additionally, we proposed two metrics, ASS and NIF, to assess the benchmarks quality and complexity, using NIF as a basis for determining the number of frames sampled by RAG-Adapter. Tests on Video-MME, MLVU, Perception Test and EgoSchema demonstrate that RAG-Adapter consistently improves accuracy across all baseline MLLMs, demonstrating our approach's simplicity and effectiveness.
\vspace{-5pt}

\paragraph{Limitations.} 
RAG-Adapter does not always retrieve all relevant frames, especially when key information is dispersed across multiple segments, often returning only a subset. Additionally, complex tasks like sentiment analysis or video summarization, which lack explicit visual cues, may further constrain its effectiveness. Moreover, the substantial preprocessing time required to encode video data into the database makes RAG-Adapter unsuitable for real-time video processing. 
While the proposed Two-Stage Retrieval and purely visual retrieval strategies mitigate this issue, future work will focus on further optimizing retrieval efficiency.
{
    \small
    \bibliographystyle{ieeenat_fullname}
    \bibliography{main}
}

\clearpage
\setcounter{page}{1}
\maketitlesupplementary

\appendix
\section*{Appendix Contents}

% \begin{itemize}
\textbf{A. Training Hyperparameters} \dotfill \pageref{sec:train}

\noindent \textbf{B. Additional Experiments} \dotfill \pageref{sec:experiments}

\textbf{B.1. Comparison of more MLLMs under different fine-tuning methods} \dotfill \pageref{sec:ablation1}

\textbf{B.2. Comparison of more MLLMs across different input frame numbers and sampling strategies} \dotfill \pageref{sec:ablation2}

\textbf{B.3. Comparison of more MLLMs under different subtitle input conditions} \dotfill \pageref{sec:ablation3}

\textbf{B.4. More Comparison Results on MLVU} \dotfill \pageref{sec:mlvu}

\noindent \textbf{C. Unreasonable Questions in the Benchmark} \dotfill \pageref{sec:error}

\noindent \textbf{D. Additional Visualization Results} \dotfill \pageref{sec:visual}

\noindent \textbf{E. Detailed NIF Statistics} \dotfill \pageref{sec:nif}

% \end{itemize}

\section{Training Hyperparameters}
\label{sec:train}
\Cref{train} provides the hyperparameters used for fine-tuning the image encoder (BGE-M3) and text encoder (CLIP-L/14) in RAG-Adapter. Both Self-supervised Contrastive Learning and Grouped-supervised Contrastive Learning use the same hyperparameter configurations.

\begin{table}[H]
% \vskip -0.15in
\caption{Training Hyperparameters for BGE-M3 and CLIP-L/14.}
    \label{train}
    % \vskip -0.15in
    \begin{center}
    % \begin{small}
    % \begin{sc}
% \resizebox{\linewidth}{!}{
\begin{tabular}{c|c|c}
\hline \hline
\multirow{2}{*}{Hyperparameter} & \multicolumn{2}{c}{Encoders} \\ \cline{2-3}
 & BGE-M3 & CLIP-L/14 \\ \hline
Batch size & 32 & 32  \\ 
Fine-tuning epochs & 2 &2  \\ 
Fine-tuning iterations & 26126&26126  \\
Temperature & 20 & 20  \\
Weight decay & 0.01 &0.01 \\ 
Learning rate & 2e-5 &1e-5 \\ 
Warm-up iterations & 2612 &2612 \\ 
Optimizer & AdamW&AdamW \\ 
Schedule &linear decay &cosine decay \\
AdamW \(\beta_{1}\) &0.9 &0.9 \\ 
AdamW \(\beta_{2}\) & 0.999&0.98 \\
AdamW \(\epsilon\) &1e-6 &1e-6 \\ \hline \hline

\end{tabular}
% }
    % \end{sc}
    % \end{small}
    \end{center}
    % \vskip -0.3in
\end{table}

\section{Additional Experiments}
\label{sec:experiments}

\subsection{Comparison of more MLLMs under different fine-tuning methods}
\label{sec:ablation1}

% In ~\Cref{evaluation benchmarks}, we mention that annotating videos requires an average of 1-2 hours. This is primarily due to RAG-Adapter's preprocessing step, where one frame per second is sampled from the video, and each frame is annotated in detail using CogVLM2, with each annotation taking approximately 5 seconds on a local NVIDIA 4090 GPU. For instance, the longest video in the Video-MME dataset is 1 hour long, resulting in a total annotation time of 5 hours. However, the actual retrieval time in the RAG-Adapter pipeline for each question is less than 10 seconds. The main purpose of using CogVLM2 is to enhance RAG-Adapter's retrieval effectiveness by using captions for each frame. To further verify the utility of captions, we conduct additional ablation experiments comparing scenarios with and without captions. In the case without captions, RAG-Adapter omits the text encoder and directly uses the image encoder and Dual Reranker to retrieve the relevant frames.

% The experimental results indicate that adding captions to each video frame enhances the retrieval performance of RAG-Adapter. This improvement arises because CLIP-L/14 primarily encodes global information from video frames, which may result in omitting detailed elements such as objects or actions. The additional encoding of captions complements the visual encoding, thereby improving the overall retrieval effectiveness of RAG-Adapter.

\begin{table}[H]
% \vskip -0.15in
\caption{Comparison under different fine-tuning methods.}
\vspace{-15pt}
    \label{experiment2_fur}
    % \vskip -0.15in
    \begin{center}
    % \begin{small}
    % \begin{sc}
\resizebox{0.75\linewidth}{!}{
\begin{tabular}{c|c|c}
\hline \hline
Models& Fine-Tuning Method &  Avg. Acc. (\%)  \\ \hline

\multirow{4}{*}{MovieChat~\cite{song2024moviechat}} & No Fine-Tuning  & 29.2 \\ \cline{2-3}
 & SCL & 28.5 (\textit{-0.7})  \\ \cline{2-3}
 & CB & 29.3 (\textit{+0.1})\\ \cline{2-3}
 & GCL &  \textbf{33.0} (\textit{+3.8}) \\ \hline

 \multirow{4}{*}{LLaMA-VID~\cite{li2025llama}} & No Fine-Tuning  & 28.9  \\ \cline{2-3}
 & SCL &  28.9 (\textit{+0.0}) \\ \cline{2-3}
 & CB & 29.6 (\textit{+0.7})\\ \cline{2-3}
 & GCL & \textbf{31.1} (\textit{+2.2}) \\ \hline

 % \multirow{4}{*}{TimeChat~\cite{ren2024timechat}} & No Fine-Tuning  & 33.7 \\ \cline{2-3}
 % & SCL & 32.6 (\textit{-1.1}) \\ \cline{2-3}
 %  &CB & 34.8 (\textit{+1.1})\\ \cline{2-3}
 % & GCL & \textbf{37.0} (\textit{+3.3}) \\ \hline

 % \multirow{4}{*}{GPT-4o~\cite{openai}} & No Fine-Tuning  & 65.9 \\ \cline{2-3}
 % & SCL & 65.2 (\textit{-0.7}) \\ \cline{2-3}
 %  & CB & 66.7 (\textit{+0.8})\\ \cline{2-3}
 % & GCL & \textbf{70.8} (\textit{+4.9}) \\ \hline \hline

\end{tabular}
}
    % \end{sc}
    % \end{small}
    \end{center}
    \vskip -0.2in
\end{table}

% \begin{table}[H]
% % \vskip -0.15in
% \caption{Comparison of accuracy for MLLMs on the Video-MME benchmark using RAG-Adapter with (w/ caps) and without (w/o caps) captions.}
%     \label{captions}
%     % \vskip -0.15in
%     \begin{center}
%     % \begin{small}
%     % \begin{sc}
% % \resizebox{\linewidth}{!}{
% \begin{tabular}{c|c|c}
% \hline \hline
% Models& Captions &  Avg. Acc. (\%)\\ \hline

% \multirow{2}{*}{MovieChat} & w/o caps & 26.3  \\ \cline{2-3}
%  & w/ caps & \textbf{33.0} (\textit{+6.7}) \\ \cline{2-3}
% \hline

%  \multirow{2}{*}{LLaMA-VID} & w/o caps & 29.3 \\ \cline{2-3}
%  & w/ caps &  \textbf{31.1} (\textit{+1.8})\\ \cline{2-3}
% \hline

%  \multirow{2}{*}{TimeChat} & w/o caps  & 31.5  \\ \cline{2-3}
%  & w/ caps  & \textbf{37.0} (\textit{+5.5})\\ \cline{2-3}
% \hline

%  \multirow{2}{*}{Chat-UniVi} & w/o caps & 34.5 \\ \cline{2-3}
%  & w/ caps & \textbf{39.7} (\textit{+5.2})\\ \cline{2-3}
% \hline \hline

% \end{tabular}
% % }
%     % \end{sc}
%     % \end{small}
%     \end{center}
%     % \vskip -0.1in
% \end{table}

\subsection{Comparison of more MLLMs across different input frame numbers and sampling strategies}
\label{sec:ablation2}

\begin{table}[H]
% \vskip -0.15in
\caption{Comparison of different sampling strategies and input frame counts.}
\vspace{-15pt}
    \label{experiment3_fur}
    % \vskip -0.15in
    \begin{center}
    % \begin{small}
    % \begin{sc}
\resizebox{0.90\linewidth}{!}{
\begin{tabular}{c|c|c|c}
\hline \hline
Models& Sampling Method & Frames Count & Avg. Acc. (\%)\\ \hline

\multirow{7}{*}{MovieChat~\cite{song2024moviechat}} 
& \multirow{4}{*}{Uniform} & 5 & 25.6  \\ \cline{3-4}
&  & 10 & 26.3  \\ \cline{3-4}
&  & 20 & 27.0 \\ \cline{3-4}
&  & 512 & 28.9\\ \cline{2-4}
& \multirow{3}{*}{RAG-Adapter} & 5 & 30.0 \\ \cline{3-4}
&  & 10 & 33.0 \\ \cline{3-4}
 &  & 20 & \textbf{33.3} \\ \hline

 \multirow{7}{*}{LLaMA-VID~\cite{li2025llama}}
 & \multirow{4}{*}{Uniform} & 5 & 26.7  \\ \cline{3-4}
 &  & 10 & 26.7  \\ \cline{3-4}
 &  & 20 & 27.1  \\ \cline{3-4}
&  & 512 & 27.8 \\ \cline{2-4}
& \multirow{3}{*}{RAG-Adapter} & 5 & 28.9 \\ \cline{3-4}
&  & 10 & \textbf{31.1} \\ \cline{3-4}
 &  & 20 & 30.8 \\ \hline

%  \multirow{7}{*}{TimeChat~\cite{ren2024timechat}}
%  & \multirow{4}{*}{Uniform} & 5 & 30.7  \\ \cline{3-4}
%  &  & 10 & 33.0  \\ \cline{3-4}
%  &  & 20 & 33.0  \\ \cline{3-4}
% &  & 64 & 32.2 \\ \cline{2-4}
% & \multirow{3}{*}{RAG-Adapter} & 5 & 36.3 \\ \cline{3-4}
% &  & 10 & \textbf{37.0} \\ \cline{3-4}
%  &  & 20 & \textbf{37.0} \\ \hline

% \multirow{7}{*}{Chat-UniVi~\cite{jin2024chat}}
% &  \multirow{4}{*}{Uniform} & 5 & 27.8  \\ \cline{3-4}
% &  & 10 & 29.6  \\ \cline{3-4}
% & & 20 & 32.6  \\ \cline{3-4}
% &  & 256 & 31.9  \\ \cline{2-4}
% & \multirow{3}{*}{RAG-Adapter} & 5 & 35.6 \\ \cline{3-4}
% &  & 10 & 39.7 \\ \cline{3-4}
%  &  & 20 & \textbf{40.0} \\ \hline \hline

\end{tabular}
}
    % \end{sc}
    % \end{small}
    \end{center}
    \vskip -0.2in
\end{table}

\subsection{Comparison of more MLLMs under different subtitle input conditions}
\label{sec:ablation3}

\begin{table}[H]
% \vskip -0.15in
\caption{Comparison between no subtitles (w/o subs), subtitles corresponding to RAG-Adapter sampled frames (w/ subs (Corresp.)), and subtitles sampled by RAG-Adapter (w/ subs (RAG-Adapter)).}
\vspace{-15pt}
    \label{experiment4_fur}
    % \vskip -0.15in
    \begin{center}
    % \begin{small}
    % \begin{sc}
\resizebox{0.9\linewidth}{!}{
\begin{tabular}{c|c|c}
\hline \hline
Models& Subtitles &  Avg. Acc. (\%)\\ \hline

\multirow{3}{*}{MovieChat~\cite{song2024moviechat}} & w/o subs & 33.0  \\ \cline{2-3}
 & w/ subs (Corresp.) &  34.1 (\textit{+1.1}) \\ \cline{2-3}
 & w/ subs (RAG-Adapter) &  \textbf{34.8} (\textit{+1.8}) \\ \hline

 \multirow{3}{*}{LLaMA-VID~\cite{li2025llama}}~\cite{li2025llama} & w/o subs & 31.1 \\ \cline{2-3}
 & w/ subs (Corresp.) &  31.9 (\textit{+0.8})\\ \cline{2-3}
 & w/ subs (RAG-Adapter) & \textbf{32.2} (\textit{+1.1})\\ \hline

 % \multirow{3}{*}{TimeChat~\cite{ren2024timechat}} & w/o subs  & 37.0 \\ \cline{2-3}
 % & w/ subs (Corresp.) & 38.2 (\textit{+1.2})\\ \cline{2-3}
 % & w/ subs (RAG-Adapter) & \textbf{39.6} (\textit{+2.6})\\ \hline

 % \multirow{3}{*}{Chat-UniVi~\cite{jin2024chat}} & w/o subs & 39.7 \\ \cline{2-3}
 % & w/ subs (Corresp.) & 40.0 (\textit{+0.3})\\ \cline{2-3}
 % & w/ subs (RAG-Adapter) & \textbf{41.5} (\textit{+1.8})\\ \hline \hline

\end{tabular}
}
    % \end{sc}
    % \end{small}
    \end{center}
    \vskip -0.3in
\end{table}

\onecolumn

\subsection{More Comparison Results on MLVU}
\label{sec:mlvu}

Due to page limitations in the main body of the paper, we have included the comprehensive experiments on MLVU mentioned in \Cref{results and analysis} in the supplementary materials. \Cref{mlvu} presents the performance of various MLLMs on the MLVU benchmark using uniform sampling versus RAG-Adapter sampling, with all models evaluated using 20 frames. The results are consistent with those on the Video-MME benchmark: RAG-Adapter sampling improves the performance of all MLLMs compared to uniform sampling. For M-Avg, GPT-4o achieves the highest improvement of 12.6\%, while VideoChat2 shows the lowest gain of 6.6\%. However, for G-Avg, the improvements are generally modest, with even a slight decline (e.g., VideoChat decreases by 0.03). This is because generative tasks require a more comprehensive understanding of the video content, meaning that the sampled frames must adequately represent the entire video. In such scenarios, RAG-Adapter sampling offers no significant advantage over uniform sampling.

\begin{table*}[h]
% \vskip -0.15in
\caption{The test results for various MLLMs on MLVU. The evaluation includes nine types of tasks: PQA (Plot QA), NQA (Needle QA), ER (Ego Reasoning), AC (Action Count), AO (Action Order), AR (Anomaly Recognition), TR (Topic Reasoning), SSC (Sub-Scene Captioning), and VS (Video Summary). ``M-Avg'' (0-100) represents the average performance across multiple-choice tasks, while ``G-Avg'' (0-10, marked by *) indicates the average performance for generative tasks.}
    \label{mlvu}
    % \vskip -0.15in
    \begin{center}
    % \begin{small}
    % \begin{sc}
\resizebox{\linewidth}{!}{
\begin{tabular}{c|c|c|c|c|c|c|c|c|c|c|c|c}
\hline \hline
\multirow{2}{*}{Models}& \multirow{2}{*}{Sampling Method} & \multicolumn{9}{c|}{Category} & \multirow{2}{*}{M-Avg} &  \multirow{2}{*}{G-Avg} \\ \cline{3-11}
&  & PQA & NQA & ER & AC & AO & AR & TR & SSC* & VS* & & \\ \hline
\multicolumn{13}{c}{\textcolor[RGB]{128,128,128}{\textit{Image MLLMs}}} \\ \hline

\multirow{2}{*}{Otter-I} & Uniform  & 31.4 & 20.8 & 34.4 & 20.0 & 40.0 & 40.0 & 40.0 & 2.15 & 1.10 & 31.8 & 1.63 \\

 & RAG-Adapter  & 34.3 (\textit{+2.9})& 54.2 (\textit{+33.4})& 46.9 (\textit{+12.5})& 10.0 (\textit{-10.0})& 50.0 (\textit{+10.0})& 40.0 (\textit{+0.0})& 53.3 (\textit{+13.3}) & 2.05 (\textit{-0.1})& 1.35 (\textit{+0.25})& 43.0 (\textit{+11.2})& 1.70 (\textit{+0.07}) \\ \hline
 
\multirow{2}{*}{LLaVA-1.6} & Uniform & 34.3 & 29.2 & 34.4 & 20.0 & 20.0 & 50.0 & 73.3 & 1.30 & 1.05 & 37.0 & 1.18 \\
 & RAG-Adapter  & 57.1 (\textit{+22.8})& 50.0 (\textit{+20.8})& 34.4 (\textit{+0.0})& 20.0 (\textit{+0.0})& 30.0 (\textit{+10.0})& 70.0 (\textit{+20.0})& 66.7 (\textit{-6.6}) & 1.95 (\textit{+0.65})& 1.30 (\textit{+0.25})& 48.1 (\textit{+11.1})& 1.63 (\textit{+0.45}) \\ \hline

% \multirow{2}{*}{GPT4-Turbo}~\cite{achiam2023gpt} &Uniform &  60.0 & \underline{71.1} &48.9 & 57.8 & 53.3 & \underline{55.6}  & 57.8  \\ 
%  &RAG-Adapter & \underline{71.1} (\textit{+11.1})& \underline{71.1} (\textit{+0.0})& 51.1 (\textit{+2.2})& 60.0 (\textit{+2.2})& 53.3 (\textit{+0.0})& \textbf{60.0} (\textit{+4.4})& 61.1 (\textit{+3.3}) \\ \hline \hline
 
\multicolumn{13}{c}{\textcolor[RGB]{128,128,128}{\textit{Video MLLMs}}} \\ \hline

\multirow{2}{*}{Otter-V} & Uniform & 28.6 & 21.7 & 18.8 & 20.0 & 40.0 & 30.0 & 33.3 & 2.20 & 1.10 & 26.1 & 1.65 \\
 & RAG-Adapter  & 31.4 (\textit{+2.8})& 37.5 (\textit{+15.8})& 28.1 (\textit{+9.3})& 40.0 (\textit{+20.0})& 40.0 (\textit{+0.0})& 40.0 (\textit{+10.0})& 40.0 (\textit{+6.7}) & 2.25 (\textit{+0.05})& 1.20 (\textit{+0.10})& 34.8 (\textit{+8.7})& 1.73 (\textit{+0.08}) \\ \hline

\multirow{2}{*}{mPlug-Owl-V} & Uniform & 25.7 & 33.3 & 37.5 & 40.0 & 20.0 &40.0 & 26.7 & 2.15 & 1.10 & 31.8 & 1.63 \\
 & RAG-Adapter  & 34.3 (\textit{+8.6})& 50.0 (\textit{+16.7})& 40.6 (\textit{+3.1})& 50.0 (\textit{+10.0})& 40.0 (\textit{+20.0})& 50.0 (\textit{+10.0})& 40.0 (\textit{+13.3}) & 2.80 (\textit{+0.65})& 1.15 (\textit{+0.05})& 42.2 (\textit{+10.4})& 1.98 (\textit{+0.35}) \\ \hline

\multirow{2}{*}{MovieChat} & Uniform & 25.7 & 29.2 & 25.0 & 30.0 & 30.0 & 20.0 & 53.3 & 1.40 & 1.05 & 29.6 & 1.23 \\
 & RAG-Adapter  & 42.9 (\textit{+17.2})& 37.5 (\textit{+8.3})& 34.4 (\textit{+9.4})& 40.0 (\textit{+10.0})& 40.0 (\textit{+10.0})& 50.0 (\textit{+30.0})& 53.3 (\textit{+0.0}) & 1.45 (\textit{+0.05})& 1.10 (\textit{+0.05})& 41.5 (\textit{+11.9})& 1.28 (\textit{+0.05}) \\ \hline

\multirow{2}{*}{VideoChat} & Uniform & 22.9 & 16.7 & 25.0 & 30.0 & 20.0 & 40.0 & 26.7 & 2.25 & 1.40 & 24.5 & 1.83 \\
 & RAG-Adapter  & 31.4 (\textit{+8.5})& 33.3 (\textit{+16.6})& 25.0 (\textit{+0.0})& 40.0 (\textit{+10.0})& 40.0 (\textit{+20.0})& 50.0 (\textit{+10.0})& 33.3 (\textit{+6.6}) & 2.30 (\textit{+0.05})& 1.30 (\textit{-0.10})& 33.3 (\textit{+8.8})& 1.80 (\textit{-0.03}) \\ \hline

\multirow{2}{*}{VideoChat2} & Uniform & 34.3 & 29.2 & 21.9 & 30.0 & 20.0 & 40.0 & 26.7 & 2.25 & 1.15 & 28.9 & 1.70 \\
 & RAG-Adapter  & 37.1 (\textit{+2.8})& 37.5 (\textit{+8.3})& 31.3 (\textit{+9.4})& 40.0 (\textit{+10.0})& 30.0 (\textit{+10.0})& 40.0 (\textit{+0.0})& 33.3 (\textit{+6.6}) & 2.65 (\textit{+0.40})& 1.20 (\textit{+0.05})& 35.5 (\textit{+6.6})& 1.93 (\textit{+0.23}) \\ \hline

\multirow{2}{*}{LLaMA-VID} & Uniform & 25.7 & 33.3 & 40.6 & 40.0 & 20.0 & 50.0 & 40.0 & 2.45 & 1.25 & 34.8 & 1.85 \\
 & RAG-Adapter  & 31.4 (\textit{+5.7})& 37.5 (\textit{+4.2})& 43.8 (\textit{+3.2})& 50.0 (\textit{+10.0})& 20.0 (\textit{+0.0})& 70.0 (\textit{+20.0})& 66.7 (\textit{+26.7}) & 2.65 (\textit{+0.20})& 1.25 (\textit{+0.00})& 43.0 (\textit{+8.2})& 1.95 (\textit{+0.10}) \\ \hline

\multirow{2}{*}{TimeChat} & Uniform & 34.3 & 41.7 & 40.6 & 40.0 & 40.0 & 20.0 & 40.0 & 1.69 & 1.10 & 37.8 & 1.40 \\
 & RAG-Adapter  & 42.9 (\textit{+8.6})& 54.2 (\textit{+12.5})& 40.6 (\textit{+0.0})& 50.0 (\textit{+10.0})& 60.0 (\textit{+20.0})& 20.0 (\textit{+0.0})& 46.7 (\textit{+6.7}) & 1.85 (\textit{+0.16})& 1.05 (\textit{-0.05})& 45.2 (\textit{+7.4})& 1.45 (\textit{+0.05}) \\ \hline

 \multirow{2}{*}{Chat-UniVi} & Uniform & 34.3 & 37.5 & 21.9 & 30.0 & 20.0 & 60.0 & 33.3 & 2.75 & 1.20 & 32.6 & 1.98 \\
 & RAG-Adapter  & 37.1 (\textit{+2.8})& 45.8 (\textit{+8.3})& 28.1 (\textit{+6.2})& 50.0 (\textit{+20.0})& 30.0 (\textit{+10.0})& 60.0 (\textit{+0.0})& 46.7 (\textit{+13.4}) & 2.95 (\textit{+0.20})& 1.20 (\textit{+0.00})& 40.0 (\textit{+7.4}) & 2.08 (\textit{+0.10}) \\ \hline

\multirow{2}{*}{GPT-4o}& Uniform & 54.3 & 54.2 & 37.5 & 40.0 & 50.0 & 50.0 & 53.3 & 1.40 & 1.55 &48.9 & 1.48 \\
 & RAG-Adapter  & 62.9 (\textit{+8.6})& 70.8 (\textit{+16.6})& 50.0 (\textit{+12.5})& 50.0 (\textit{+10.0})& 60.0 (\textit{+10.0})& 80.0 (\textit{+30.0})& 60.0 (\textit{+6.7}) & 2.35 (\textit{+0.95})& 1.85 (\textit{+0.30})& 61.5 (\textit{+12.6})& 2.10 (\textit{+0.62}) \\ \hline \hline
\end{tabular}
}
    % \end{sc}
    % \end{small}
    \end{center}
    % \vskip -0.2in
\end{table*}

\newpage

\section{Unreasonable Questions in the Benchmark}
\label{sec:error}
\begin{figure*}[h]
    % \vskip -0.1in
    \begin{center}
    \centerline{\includegraphics[width=1\linewidth]{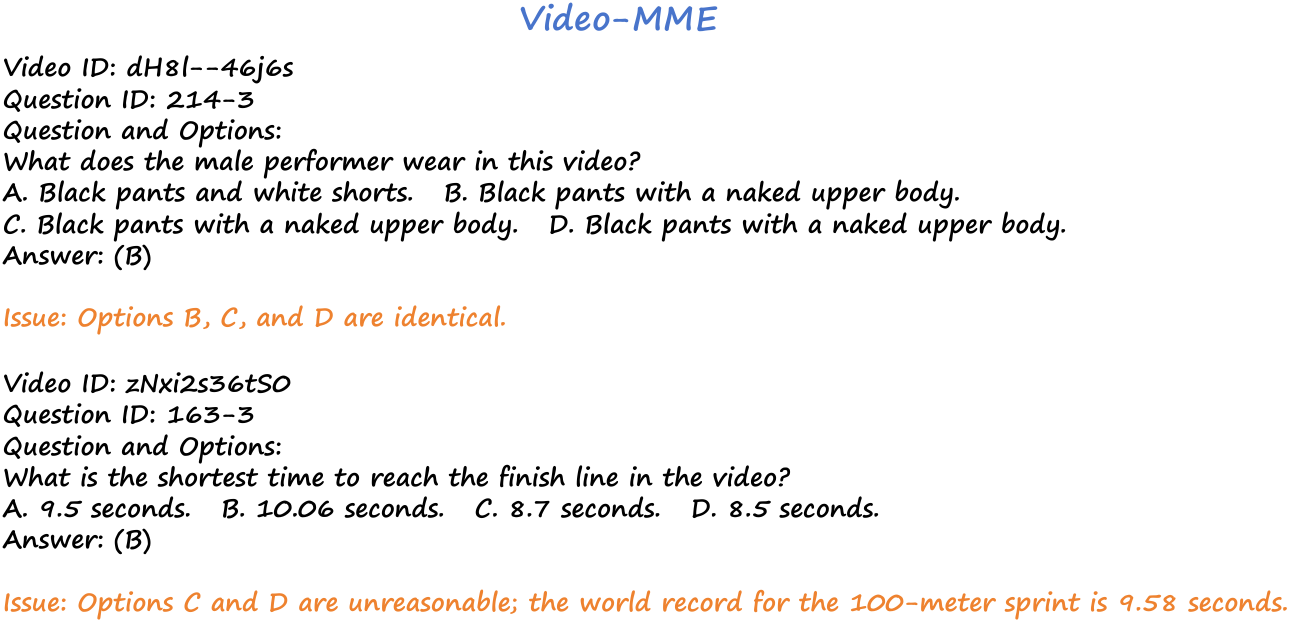}}
     \caption{Issues identified in Video-MME during NIF calculations using RAG-Adapter.}
     \label{issue_video_mme}
     \end{center}
     \vskip -0.3in
\end{figure*}

\begin{figure*}[h]
    % \vskip -0.1in
    \begin{center}
    \centerline{\includegraphics[width=1\linewidth]{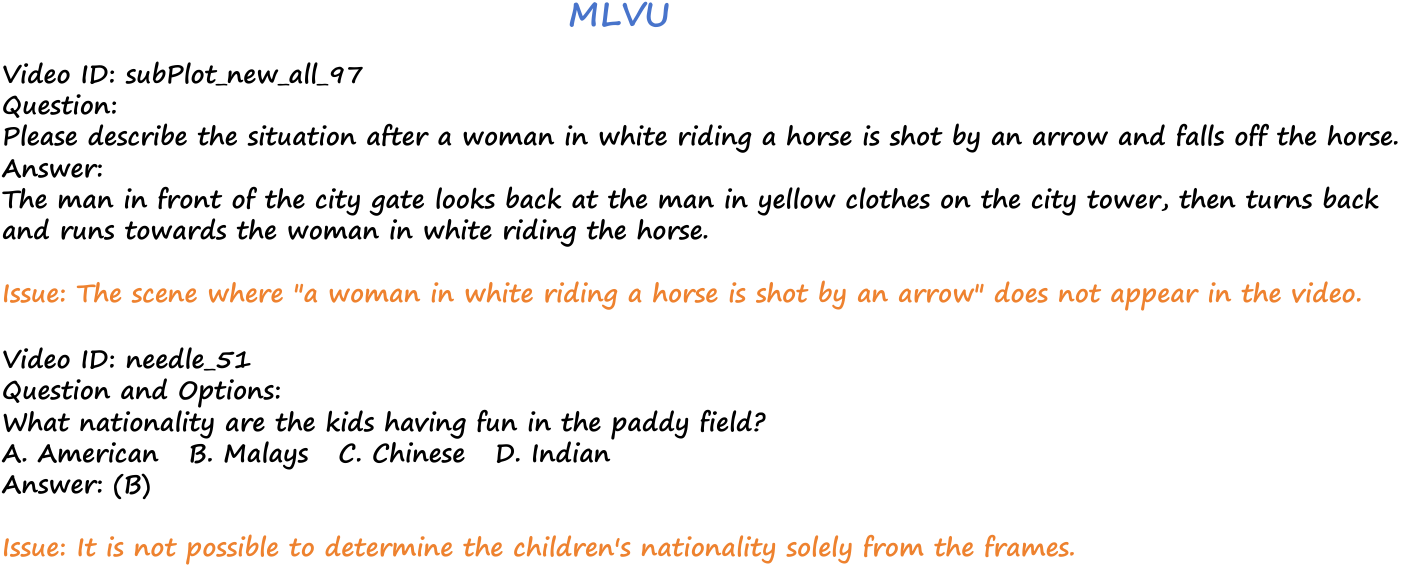}}
     \caption{Issues identified in MLVU during NIF calculations using RAG-Adapter.}
     \label{issue_mlvu}
     \end{center}
     \vskip -0.3in
\end{figure*}

\begin{figure*}[h]
    % \vskip -0.1in
    \begin{center}
    \centerline{\includegraphics[width=1\linewidth]{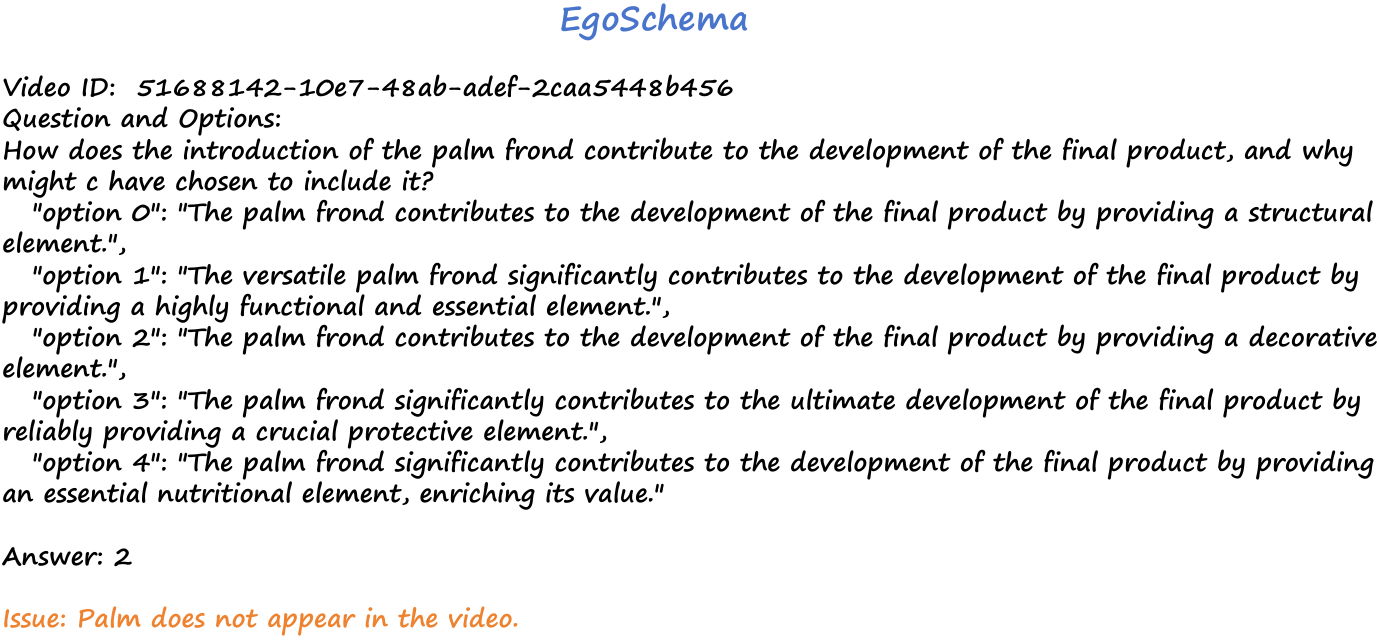}}
     \caption{Issue 1 identified in EgoSchema during NIF calculations using RAG-Adapter.}
     \label{issue_egoschema1}
     \end{center}
     \vskip -0.3in
\end{figure*}

\begin{figure*}[h]
    % \vskip -0.1in
    \begin{center}
    \centerline{\includegraphics[width=1\linewidth]{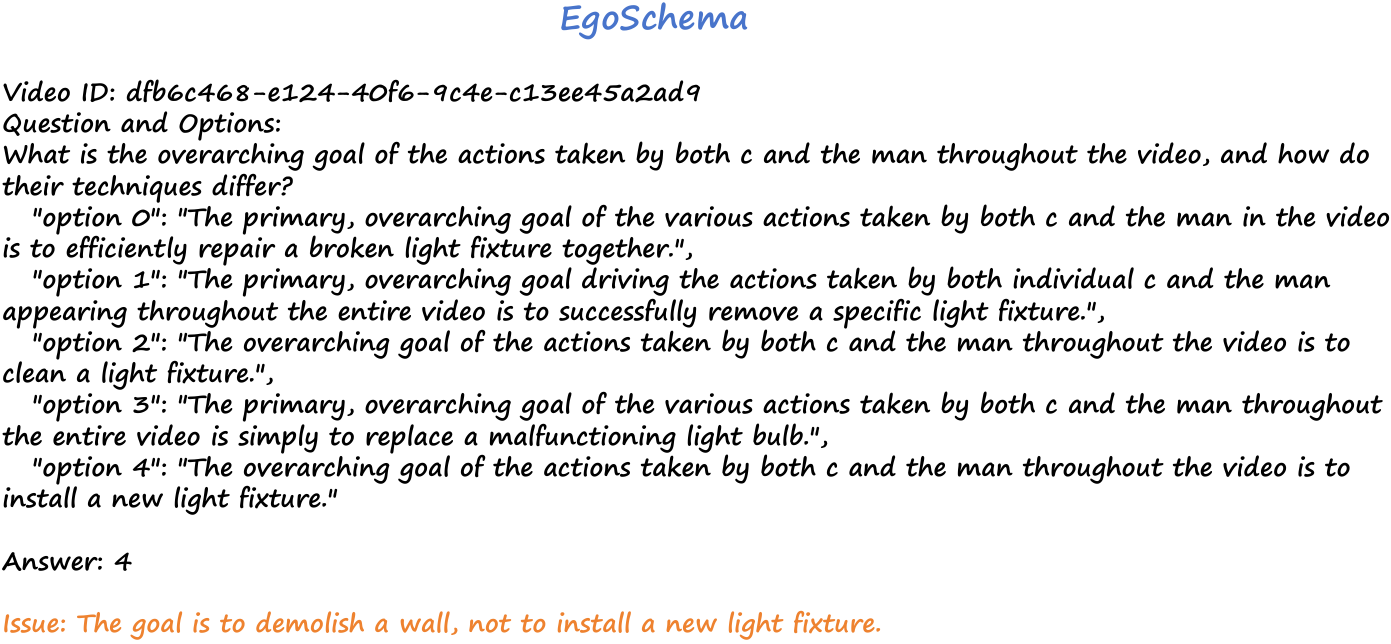}}
     \caption{Issue 2 identified in EgoSchema during NIF calculations using RAG-Adapter.}
     \label{issue_egoschema2}
     \end{center}
     \vskip -0.3in
\end{figure*}

\begin{figure*}[h]
    % \vskip -0.1in
    \begin{center}
    \centerline{\includegraphics[width=1\linewidth]{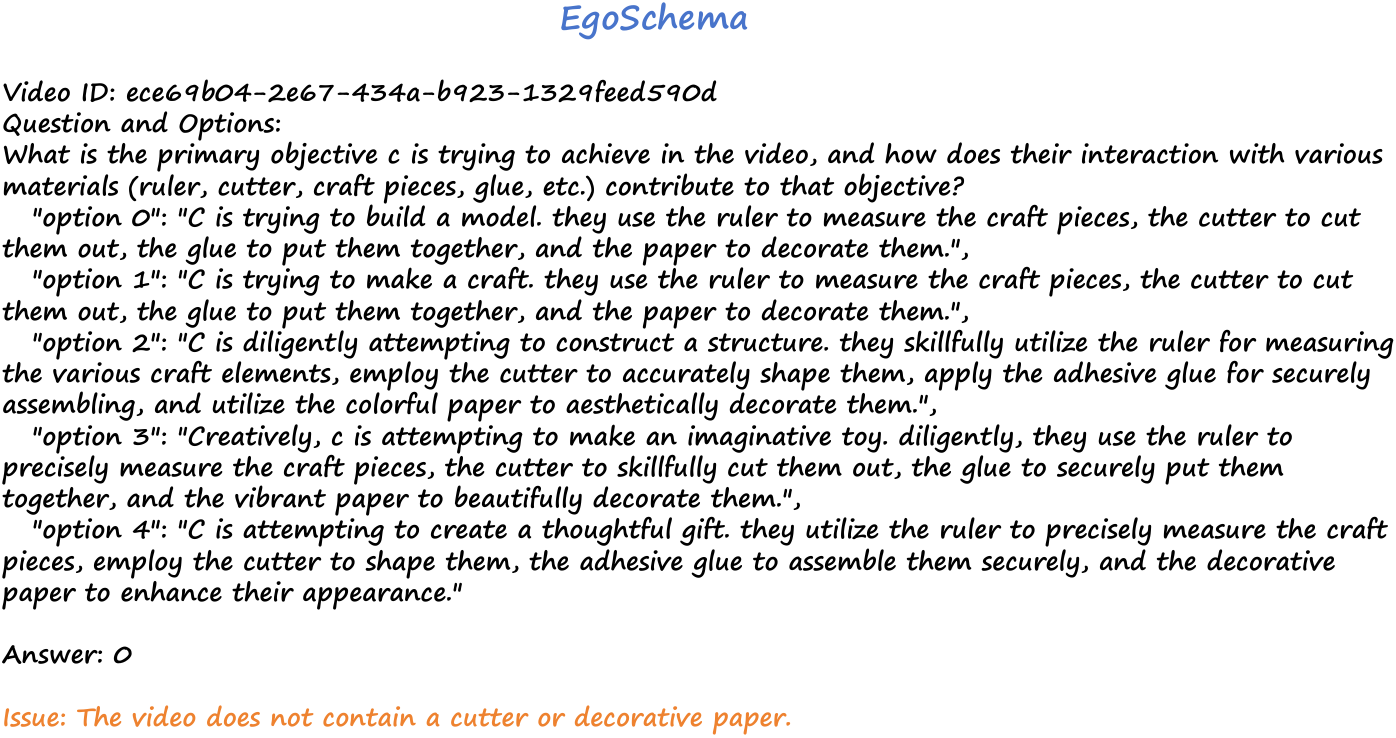}}
     \caption{Issue 3 identified in EgoSchema during NIF calculations using RAG-Adapter.}
     \label{issue_egoschema3}
     \end{center}
     \vskip -0.3in
\end{figure*}

\begin{figure*}[h]
    % \vskip -0.1in
    \begin{center}
    \centerline{\includegraphics[width=1\linewidth]{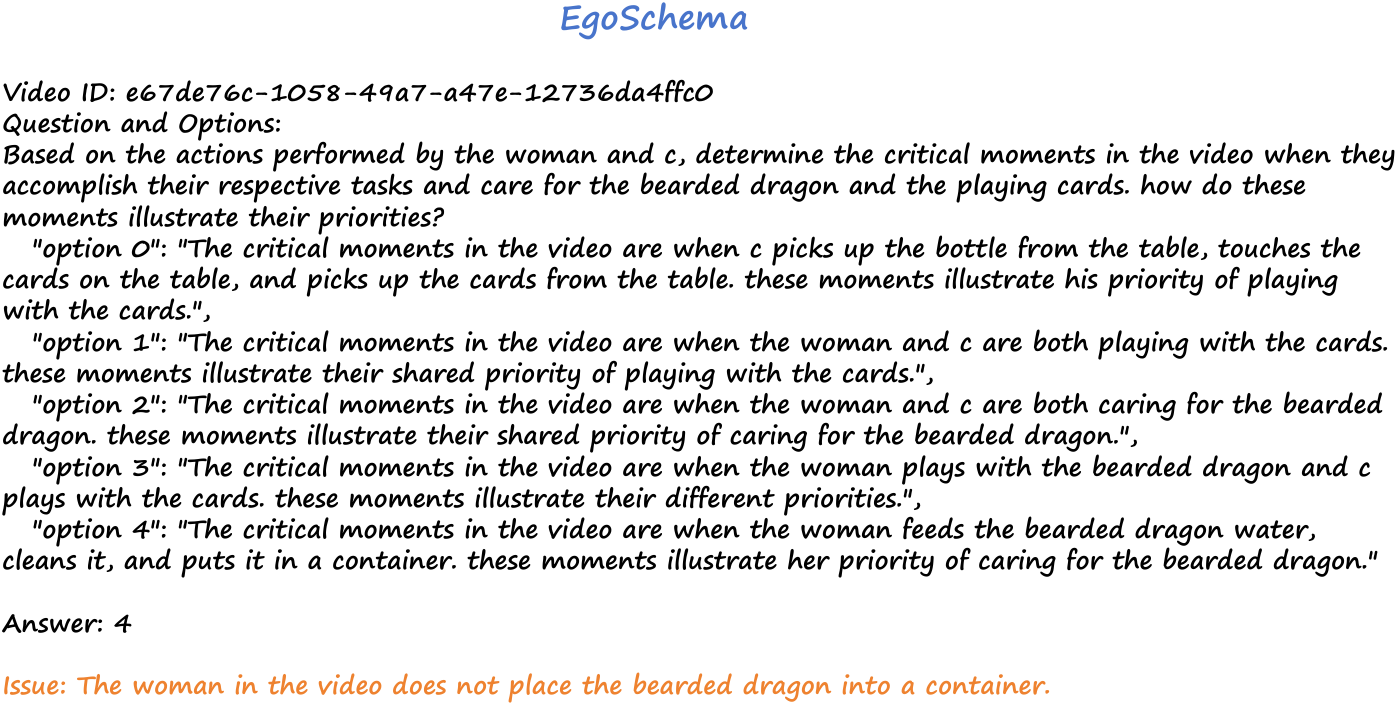}}
     \caption{Issue 4 identified in EgoSchema during NIF calculations using RAG-Adapter.}
     \label{issue_egoschema4}
     \end{center}
     \vskip -0.3in
\end{figure*}

While using RAG-Adapter to assist in calculating the NIF values for benchmarks, we identified a few unreasonable questions, detailed in \Cref{issue_video_mme,issue_mlvu,issue_egoschema1,issue_egoschema2,issue_egoschema3,issue_egoschema4}. Despite these issues, both benchmarks maintain high overall quality. This suggests that RAG-Adapter is an effective tool for evaluating and refining long video benchmarks, significantly reducing manual verification effort and enhancing benchmark quality.

\onecolumn
\newpage

\section{Additional Visualization Results}
\label{sec:visual}

\begin{figure*}[h]
    % \vskip -0.1in
    \begin{center}
    \centerline{\includegraphics[width=1\linewidth]{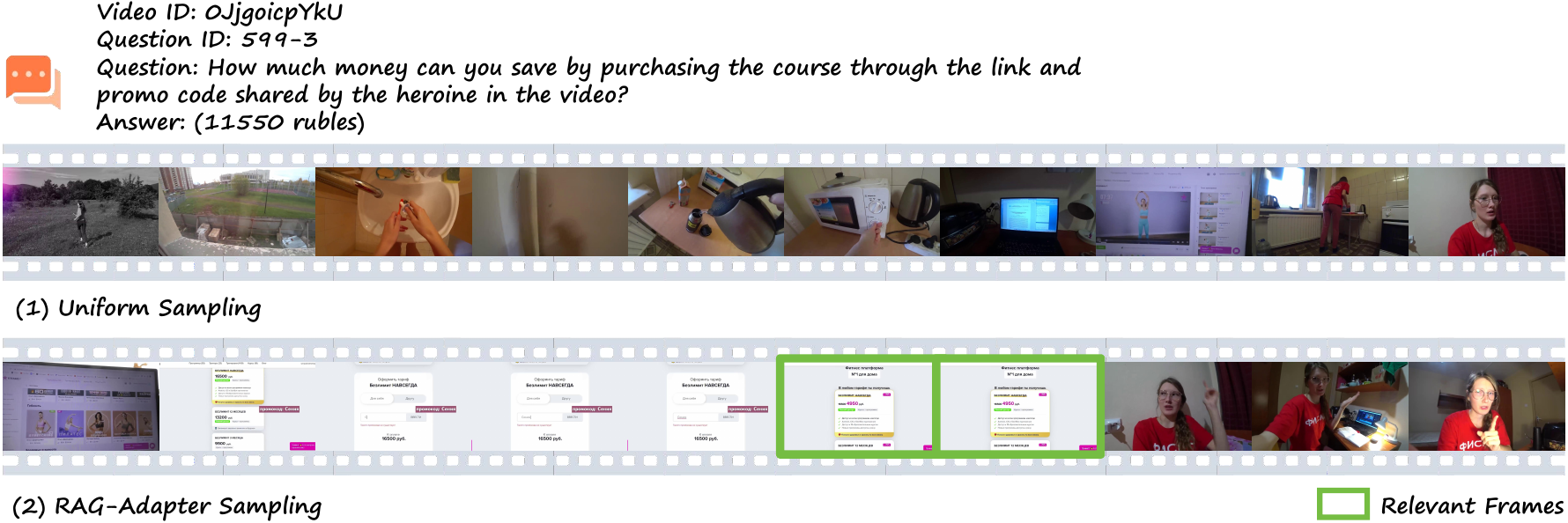}}
     \caption{Comparison of RAG-Adapter and uniform sampling results: The answer to the question appears at the 527th and 528th seconds of the video, showing the original course price as 16,500 rubles and the current price as 4,950 rubles, resulting in a total saving of 11,550 rubles. RAG-Adapter accurately identifies these two consecutive key frames, while uniform sampling tends to miss them.}
     \label{more_visual1}
     \end{center}
     \vskip -0.3in
\end{figure*}

\begin{figure*}[h]
    % \vskip -0.1in
    \begin{center}
    \centerline{\includegraphics[width=1\linewidth]{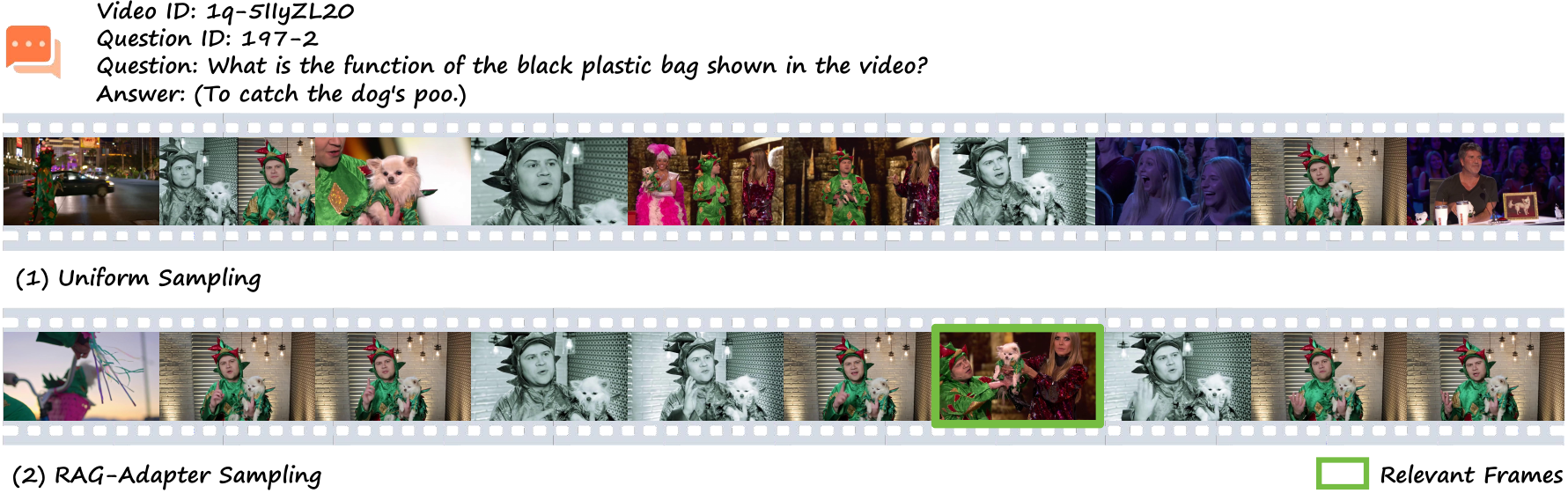}}
     \caption{Comparison of RAG-Adapter and uniform sampling results: The answer to the question appears between the 68th and 73rd seconds of the video, where a man picks up a black plastic bag to clean up after his dog. RAG-Adapter accurately identifies one key frame depicting this action (the rest of the frames contain highly similar content), whereas uniform sampling misses it.}
     \label{more_visual2}
     \end{center}
     \vskip -0.3in
\end{figure*}

\begin{figure*}[h]
    % \vskip -0.1in
    \begin{center}
    \centerline{\includegraphics[width=1\linewidth]{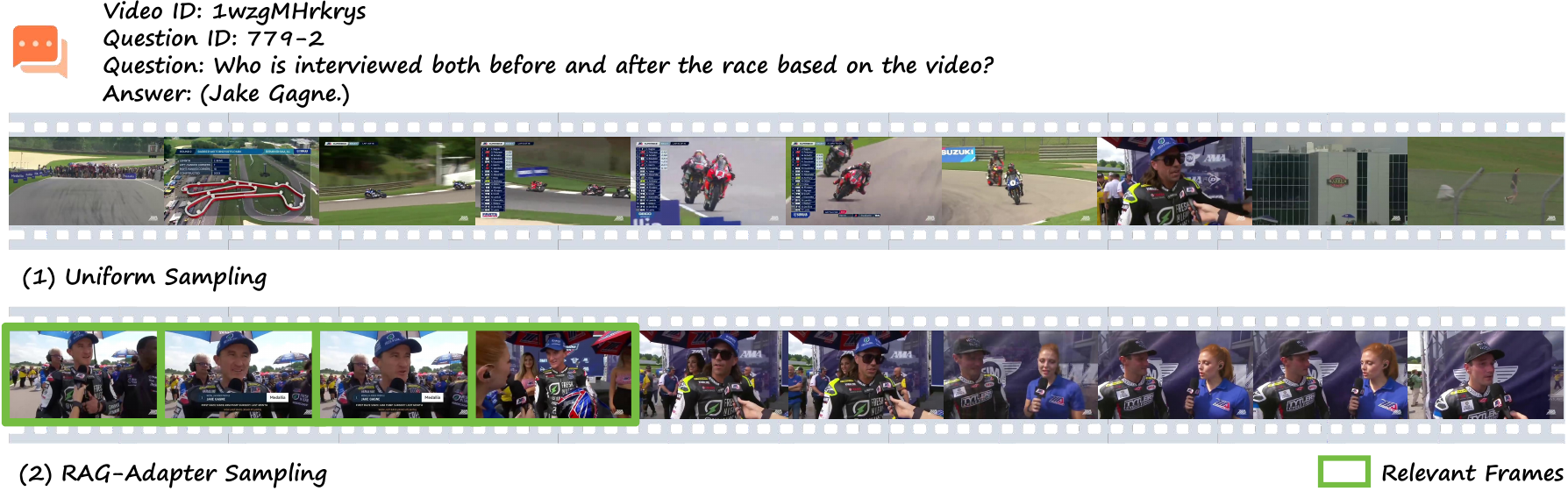}}
     \caption{Comparison of RAG-Adapter and uniform sampling results: The answer to the question appears between 270-294 seconds and 2329-2414 seconds of the video, where a reporter interviews Jake Gagne both before and after the race. RAG-Adapter accurately identifies frames at 280s, 284s, and 285s before the race, and 2334s after the race, showing the interview with Jake Gagne. The frames at 284s and 285s explicitly display Jake Gagne's name. In contrast, uniform sampling only captures a frame at 2428s, which shows an interview with another competitor, missing the key moments relevant to Jake Gagne.}
     \label{more_visual3}
     \end{center}
     \vskip -0.3in
\end{figure*}

\begin{figure*}[h]
    % \vskip -0.1in
    \begin{center}
    \centerline{\includegraphics[width=1\linewidth]{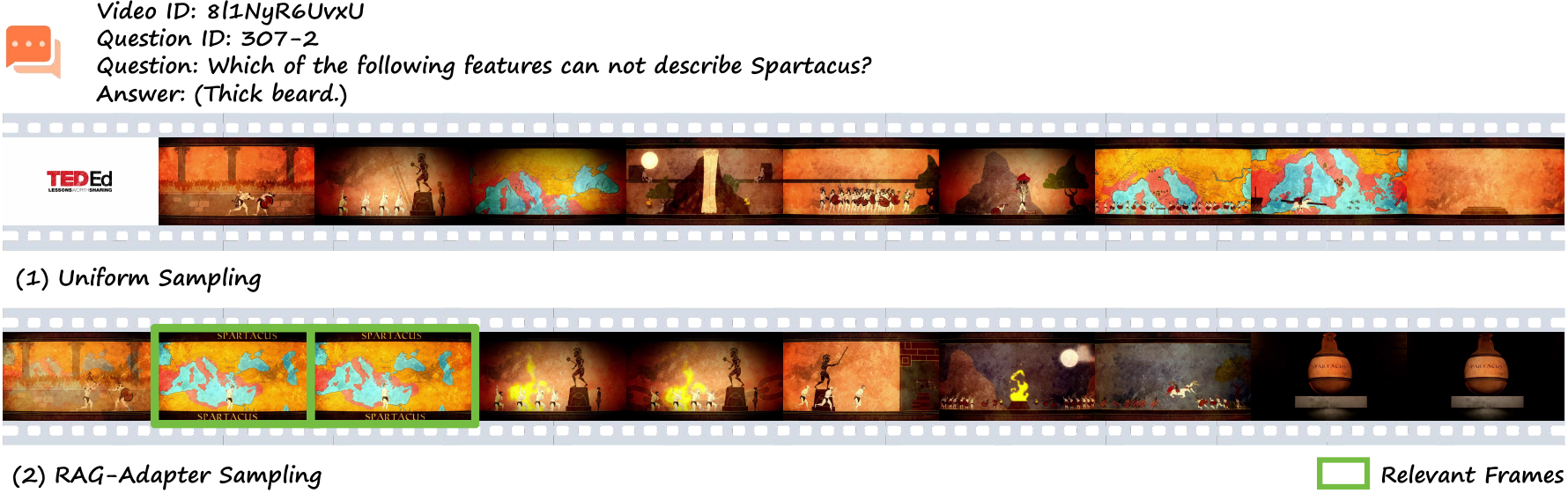}}
     \caption{Comparison of RAG-Adapter and uniform sampling results: The answer to the question appears at the 39s and 40s of the video, clearly displaying the name Spartacus and his appearance (notably without a thick beard). RAG-Adapter accurately identifies these two key frames, whereas uniform sampling fails to capture them, missing essential visual details.}
     \label{more_visual4}
     \end{center}
     \vskip -0.3in
\end{figure*}

\begin{figure*}[h]
    % \vskip -0.1in
    \begin{center}
    \centerline{\includegraphics[width=1\linewidth]{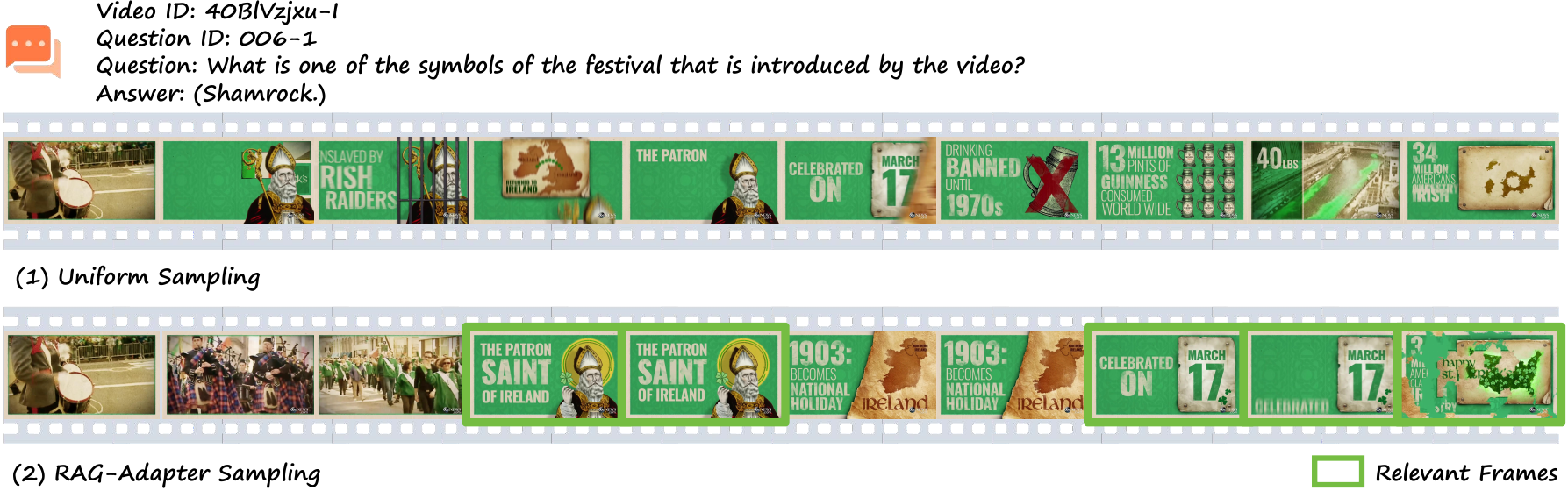}}
     \caption{Comparison of RAG-Adapter and uniform sampling results: The Shamrock logo appears in the video at 7-8 seconds, 23-25 seconds, 42-44 seconds, 53-55 seconds, and 95-105 seconds. Despite its frequent appearance, uniform sampling fails to capture any of these key frames, whereas RAG-Adapter successfully identifies key frames at 42-43 seconds, 54-55 seconds, and the 101st second.}
     \label{more_visual5}
     \end{center}
     \vskip -0.3in
\end{figure*}

\Cref{more_visual1,more_visual2,more_visual3,more_visual4,more_visual5} present additional comparisons between uniform sampling of 10 frames and RAG-Adapter sampling of 10 frames on the Video-MME benchmark. The results demonstrate that RAG-Adapter more accurately identifies video frames relevant to the question. In contrast, uniform sampling often misses these critical frames, resulting in MLLMs lacking essential information when answering questions.

\onecolumn
\newpage
\section{Detailed NIF Statistics}
\label{sec:nif}

\begin{figure*}[h]
    % \vskip -0.1in
    \begin{center}
    \centerline{\includegraphics[width=0.6\linewidth]{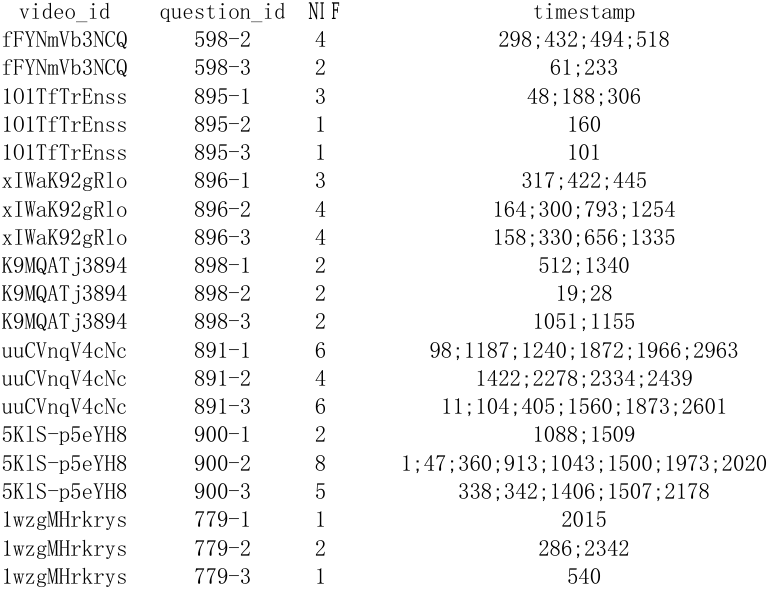}}
     \caption{The NIF values for each question in Video-MME.}
     \label{video_mme_nif6}
     \end{center}
     \vskip -0.4in
\end{figure*}

\begin{figure*}[t]
    % \vskip -0.1in
    \begin{center}
    \centerline{\includegraphics[width=0.6\linewidth]{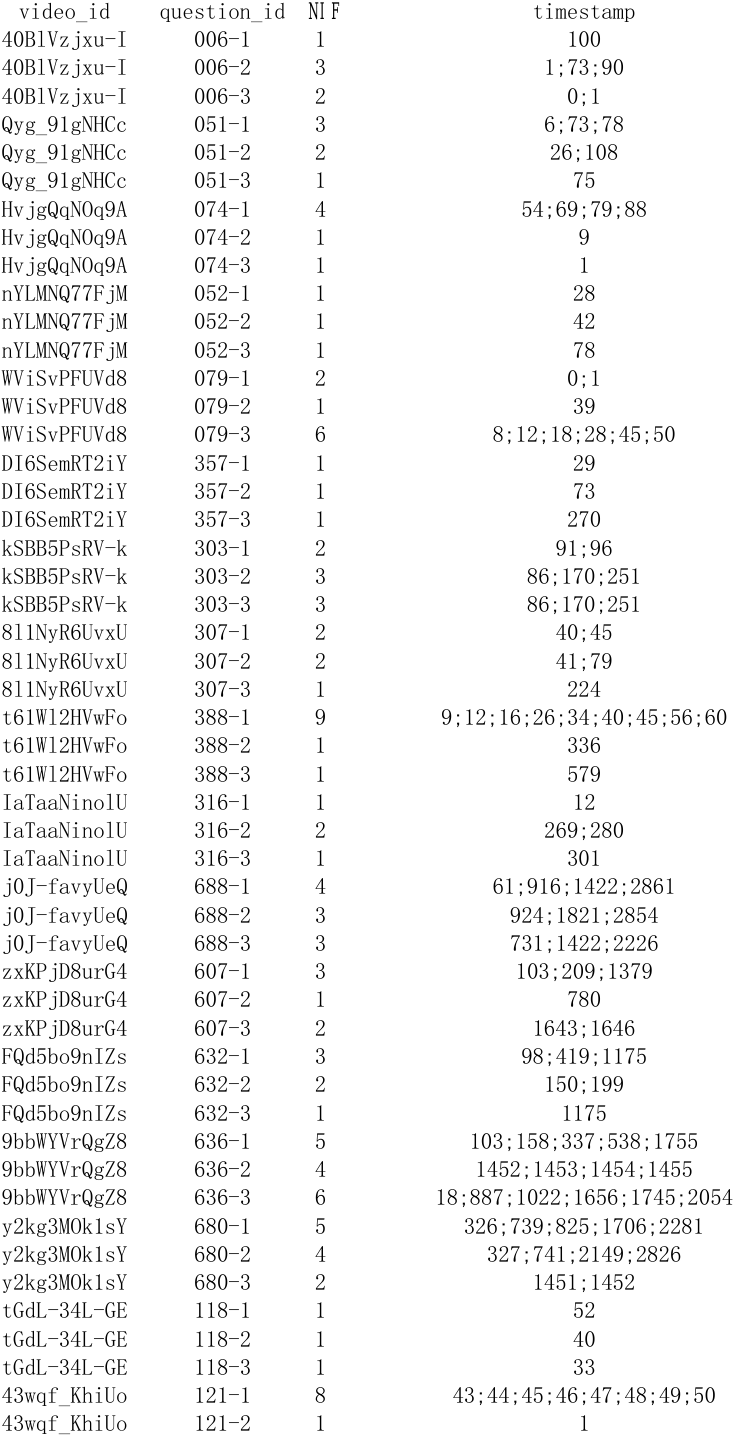}}
     \caption{The NIF values for each question in Video-MME.}
     \label{video_mme_nif1}
     \end{center}
     \vskip -0.4in
\end{figure*}

\begin{figure*}[t]
    % \vskip -0.1in
    \begin{center}
    \centerline{\includegraphics[width=0.6\linewidth]{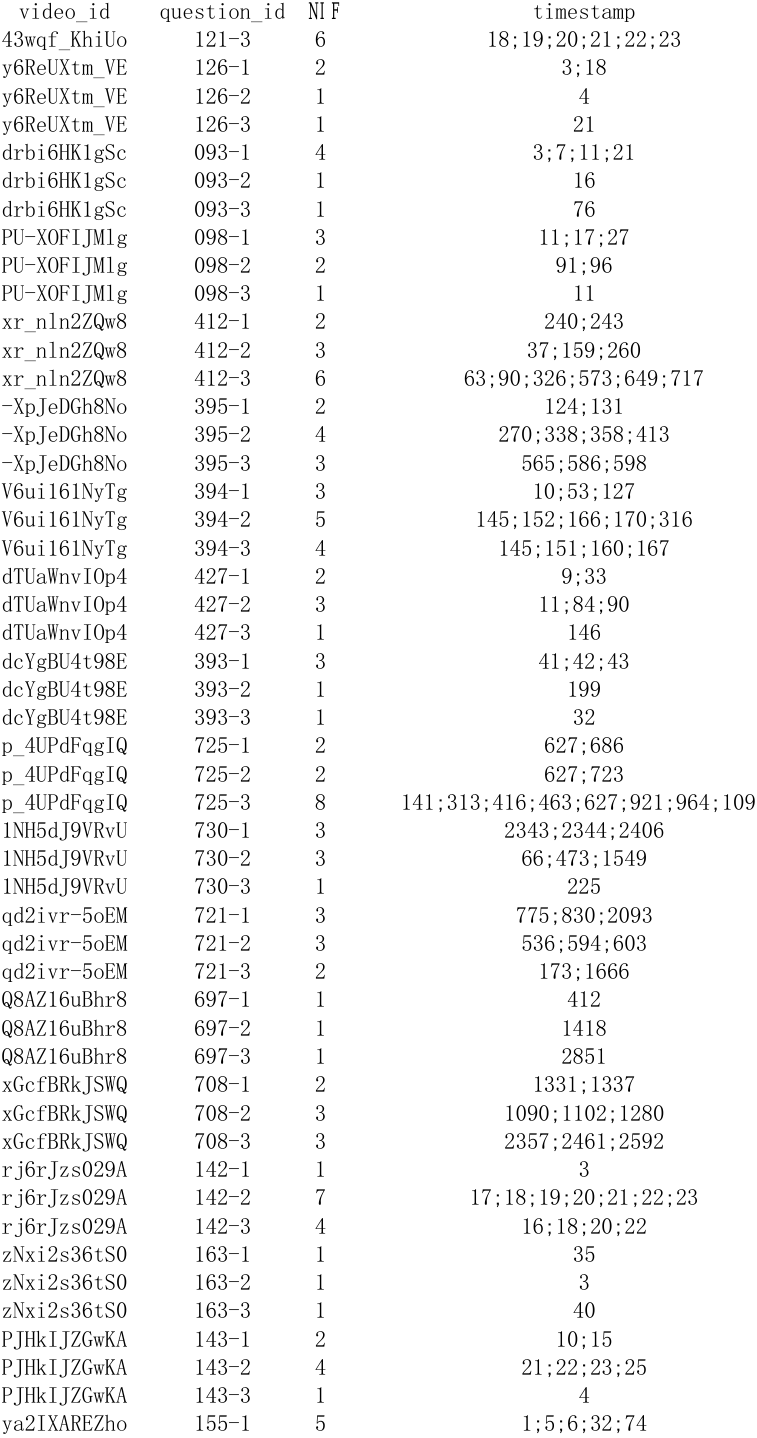}}
     \caption{The NIF values for each question in Video-MME.}
     \label{video_mme_nif2}
     \end{center}
     \vskip -0.4in
\end{figure*}

\begin{figure*}[t]
    % \vskip -0.1in
    \begin{center}
    \centerline{\includegraphics[width=0.6\linewidth]{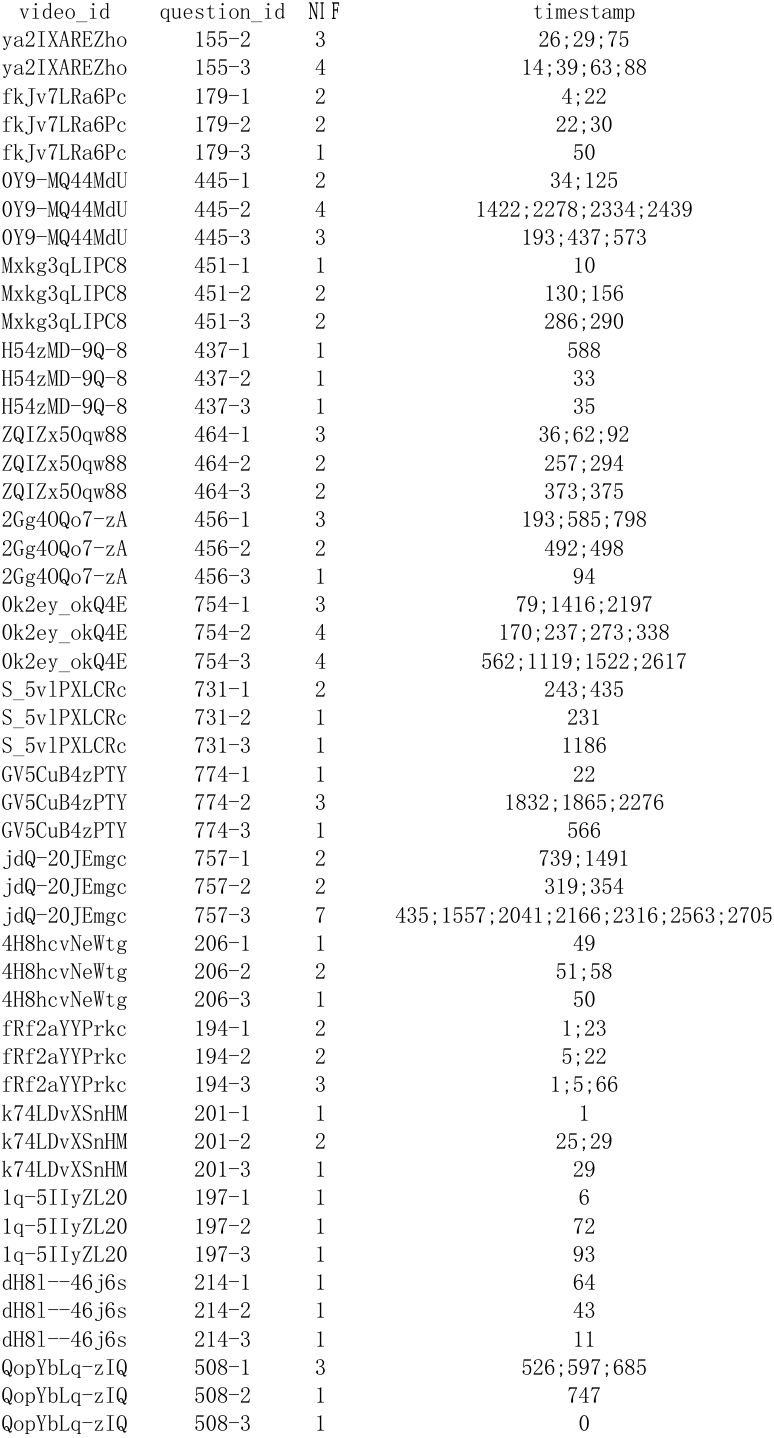}}
     \caption{The NIF values for each question in Video-MME.}
     \label{video_mme_nif3}
     \end{center}
     \vskip -0.4in
\end{figure*}

\begin{figure*}[t]
    % \vskip -0.1in
    \begin{center}
    \centerline{\includegraphics[width=0.6\linewidth]{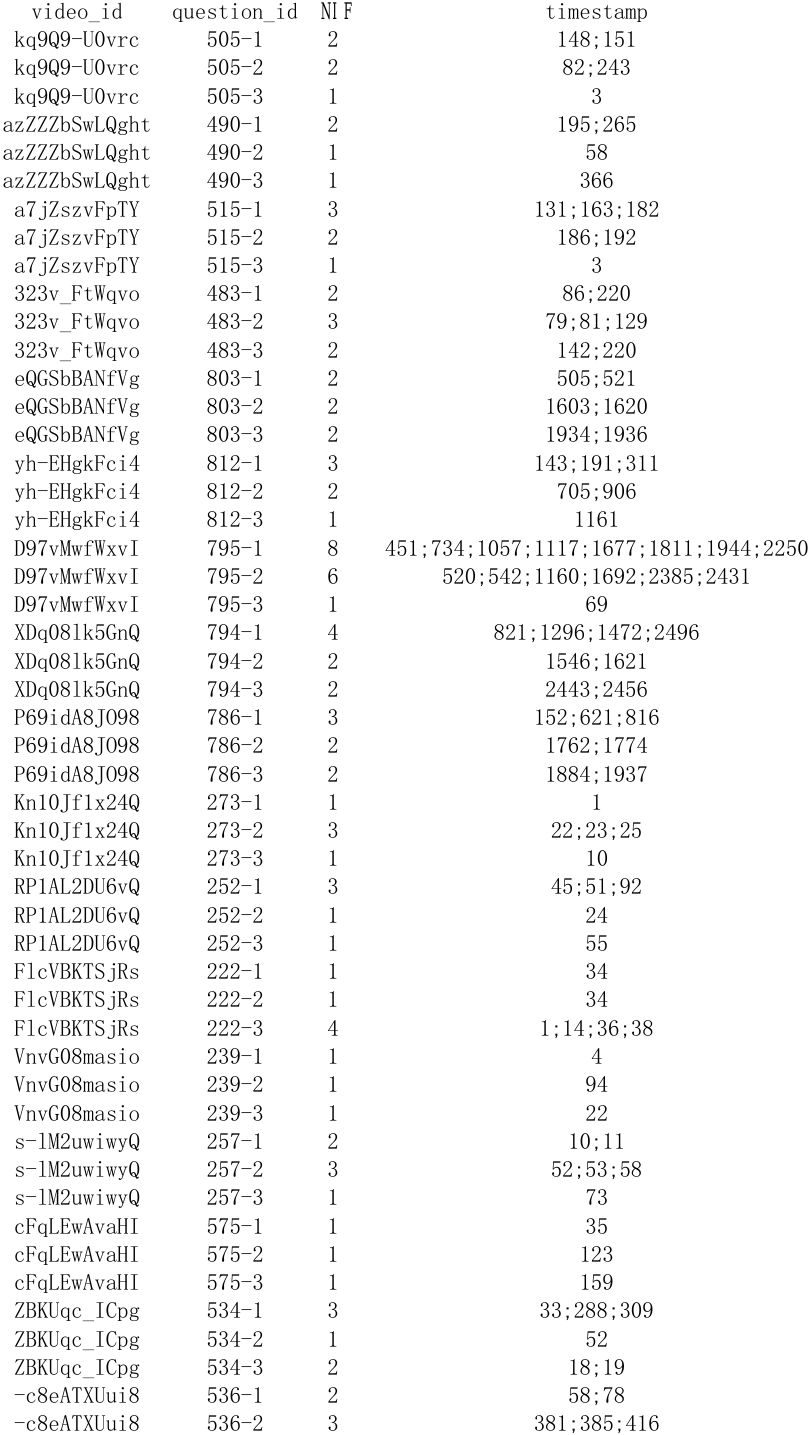}}
     \caption{The NIF values for each question in Video-MME.}
     \label{video_mme_nif4}
     \end{center}
     \vskip -0.4in
\end{figure*}

\begin{figure*}[t]
    % \vskip -0.1in
    \begin{center}
    \centerline{\includegraphics[width=0.6\linewidth]{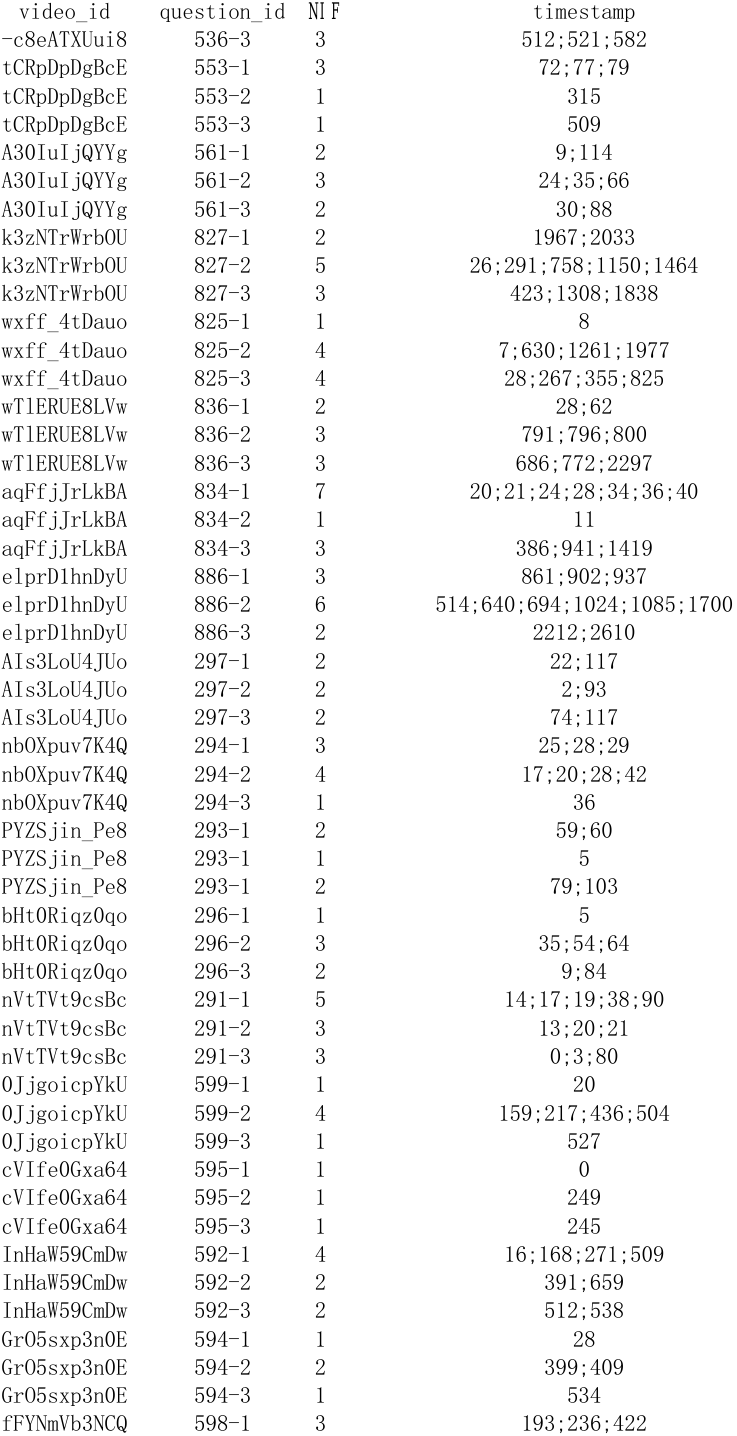}}
     \caption{The NIF values for each question in Video-MME.}
     \label{video_mme_nif5}
     \end{center}
     \vskip -0.4in
\end{figure*}

\begin{figure*}[t]
    % \vskip -0.1in
    \begin{center}
    \centerline{\includegraphics[width=0.7\linewidth]{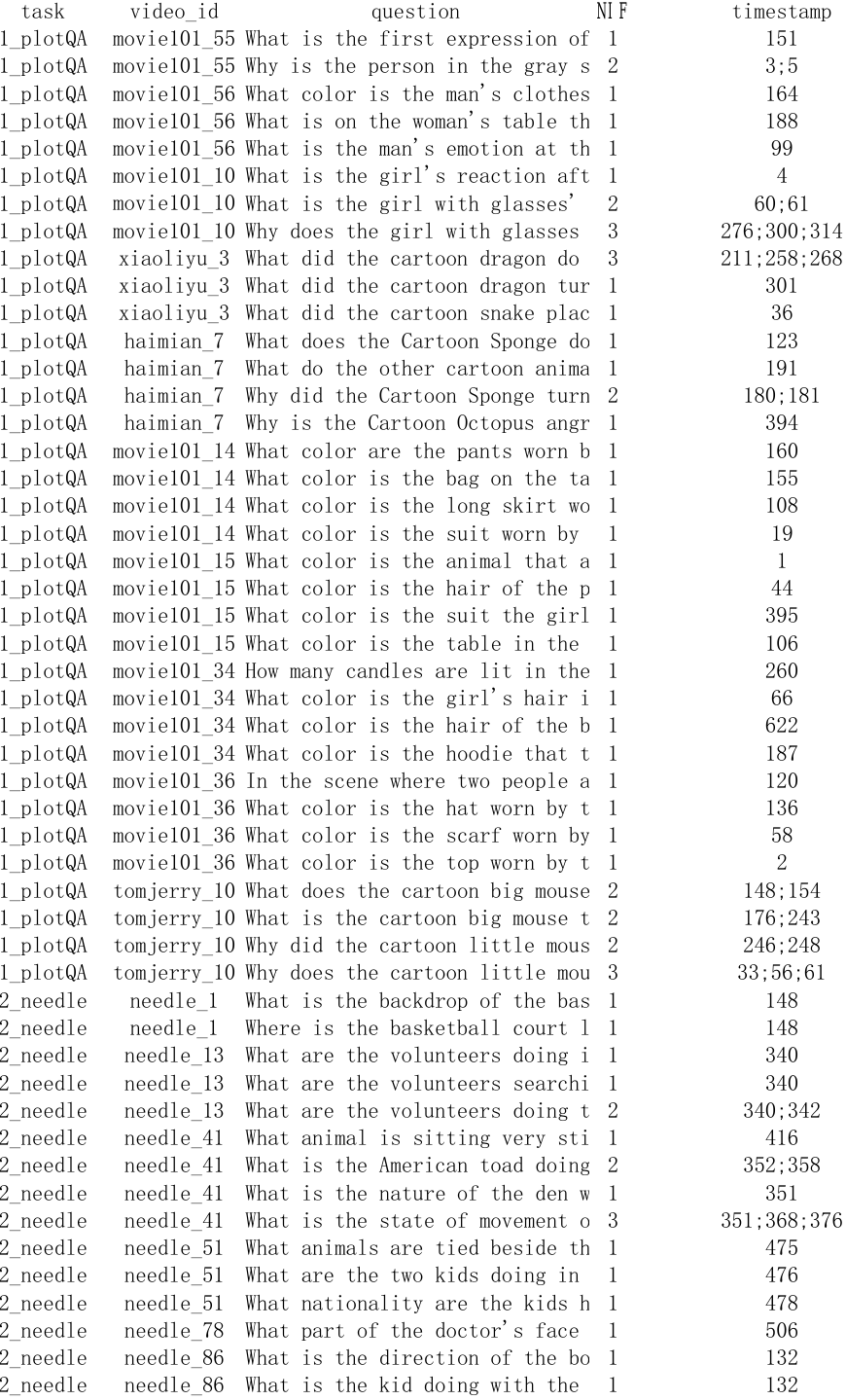}}
     \caption{The NIF values for each question in MLVU.}
     \label{mlvu_nif1}
     \end{center}
     \vskip -0.4in
\end{figure*}

\begin{figure*}[t]
    % \vskip -0.1in
    \begin{center}
    \centerline{\includegraphics[width=0.7\linewidth]{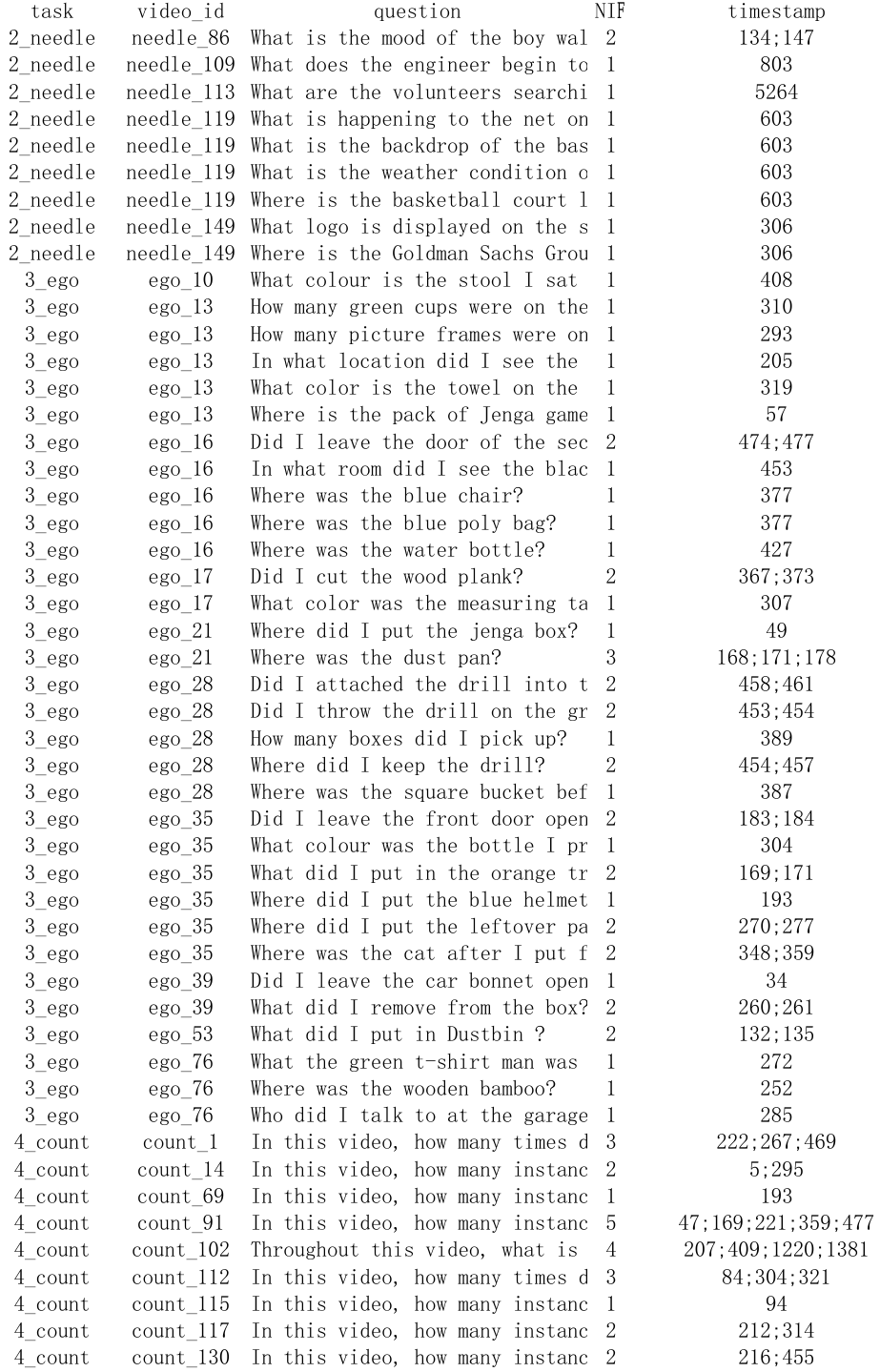}}
     \caption{The NIF values for each question in MLVU.}
     \label{mlvu_nif2}
     \end{center}
     \vskip -0.4in
\end{figure*}

\begin{figure*}[t]
    % \vskip -0.1in
    \begin{center}
    \centerline{\includegraphics[width=0.7\linewidth]{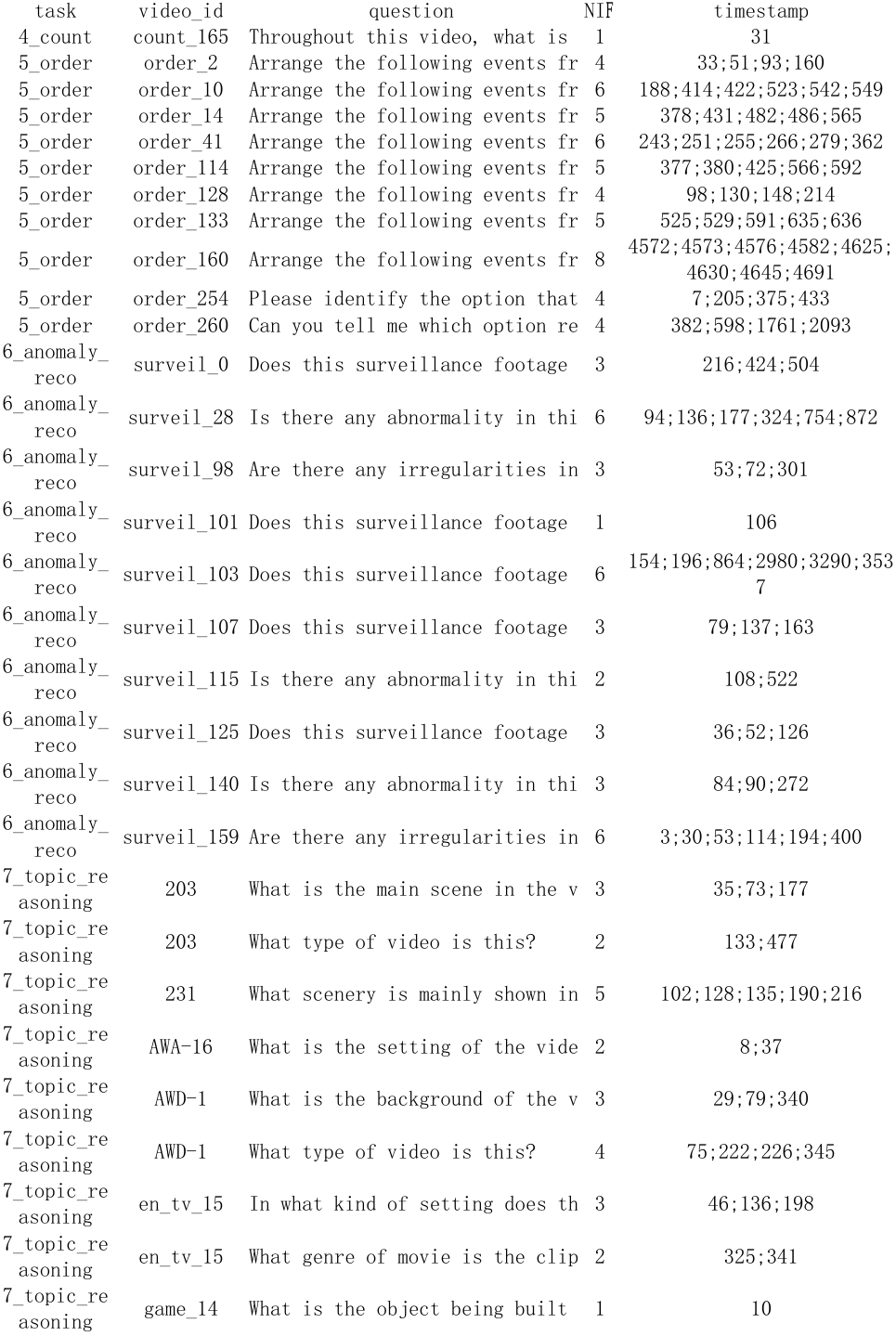}}
     \caption{The NIF values for each question in MLVU.}
     \label{mlvu_nif3}
     \end{center}
     \vskip -0.3in
\end{figure*}

\begin{figure*}[t]
    % \vskip -0.1in
    \begin{center}
    \centerline{\includegraphics[width=0.7\linewidth]{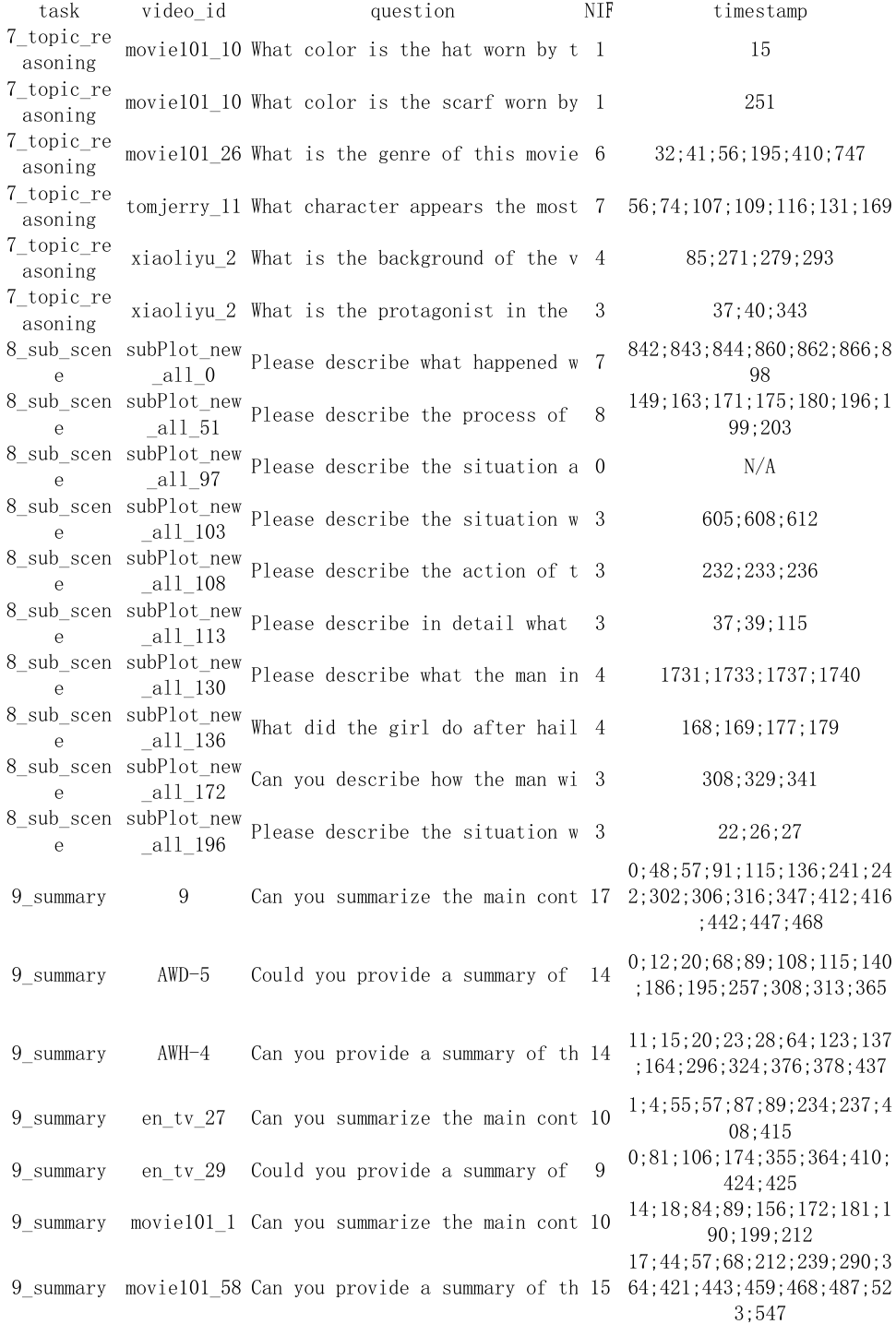}}
     \caption{The NIF values for each question in MLVU.}
     \label{mlvu_nif4}
     \end{center}
     \vskip -0.3in
\end{figure*}

\begin{figure*}[t]
    % \vskip -0.1in
    \begin{center}
    \centerline{\includegraphics[width=0.7\linewidth]{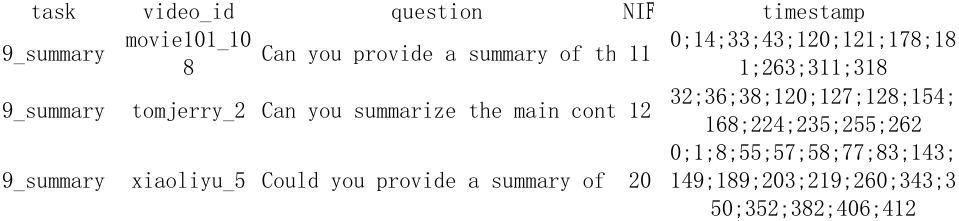}}
     \caption{The NIF values for each question in MLVU.}
     \label{mlvu_nif5}
     \end{center}
     \vskip -0.3in
\end{figure*}

\begin{figure*}[t]
    % \vskip -0.1in
    \begin{center}
    \centerline{\includegraphics[width=0.55\linewidth]{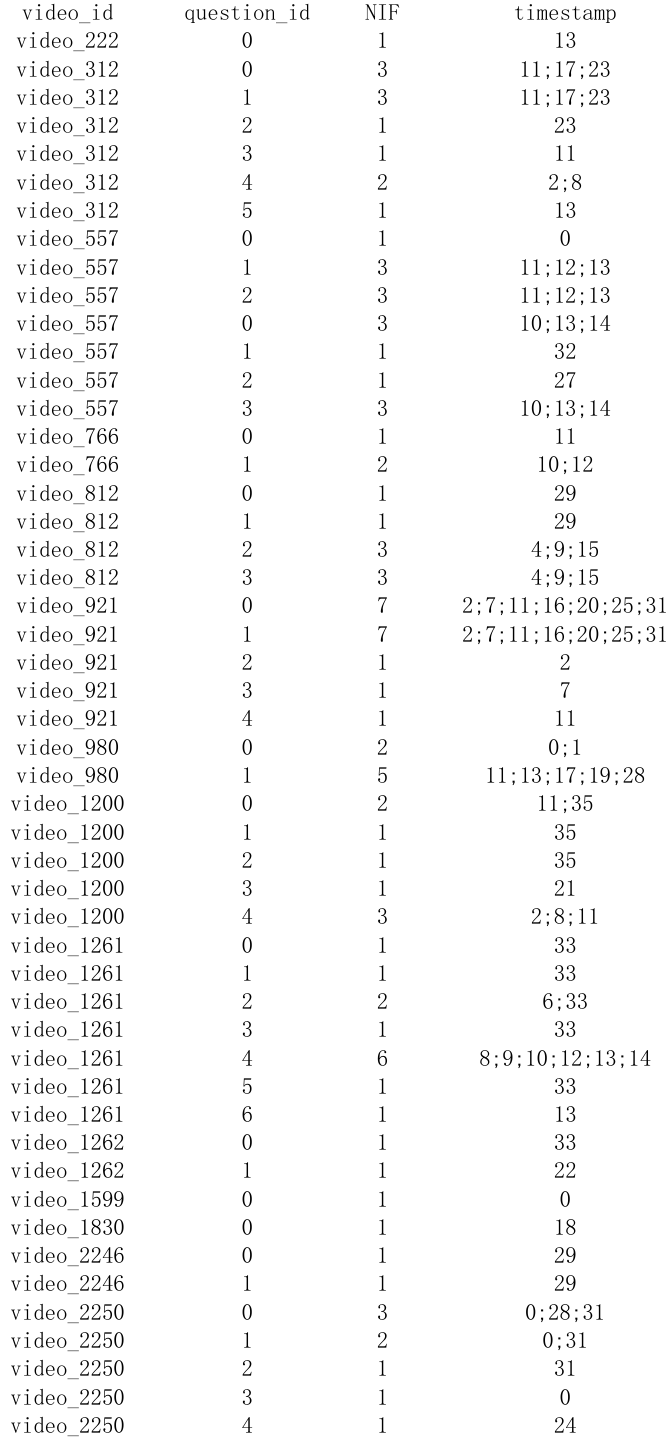}}
     \caption{The NIF values for each question in Perception Test.}
     \label{perception_test_nif1}
     \end{center}
     \vskip -0.3in
\end{figure*}

\begin{figure*}[t]
    % \vskip -0.1in
    \begin{center}
    \centerline{\includegraphics[width=0.55\linewidth]{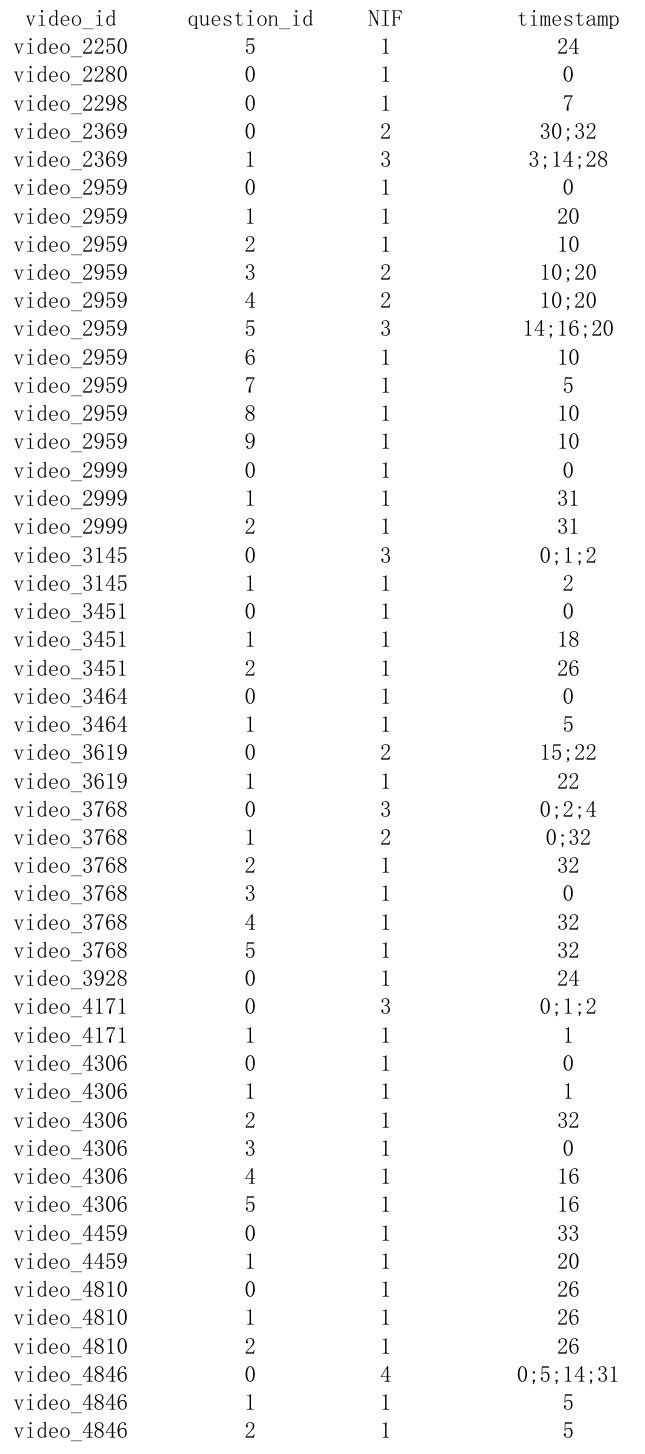}}
     \caption{The NIF values for each question in Perception Test.}
     \label{perception_test_nif2}
     \end{center}
     \vskip -0.3in
\end{figure*}

\begin{figure*}[t]
    % \vskip -0.1in
    \begin{center}
    \centerline{\includegraphics[width=0.55\linewidth]{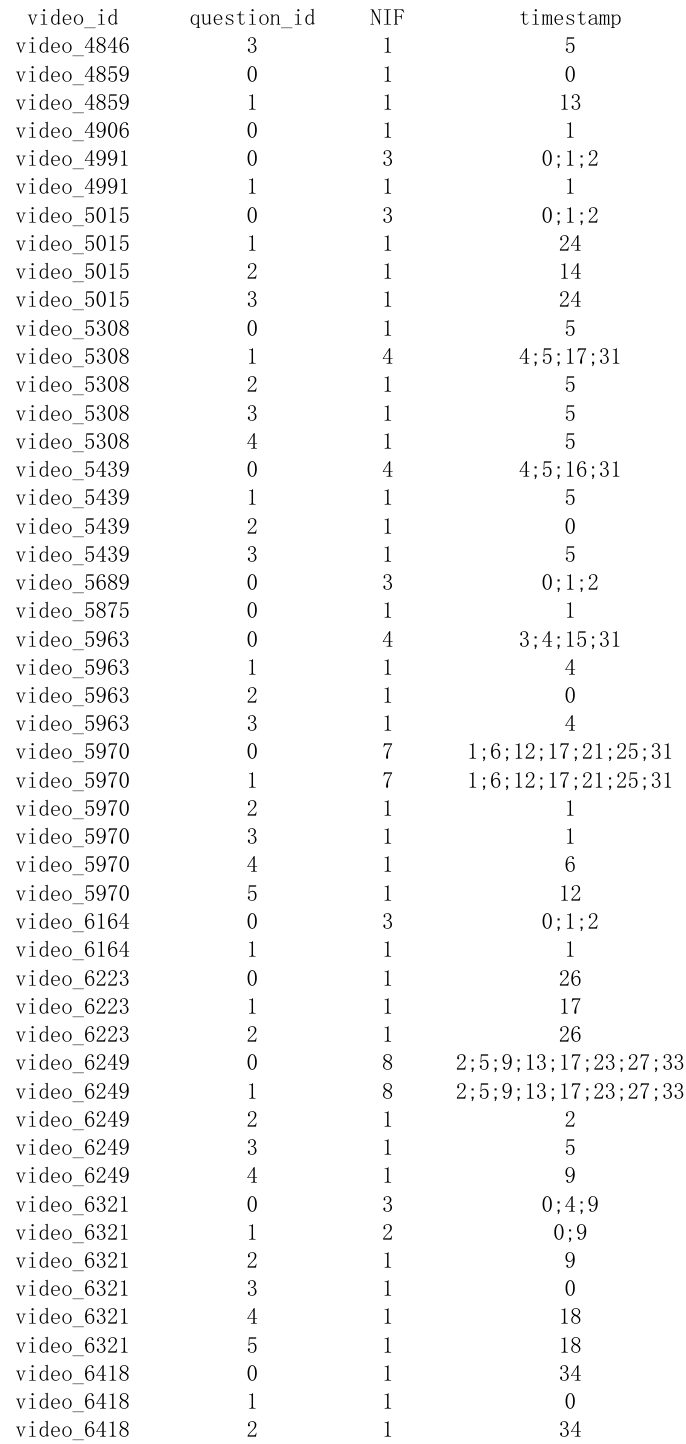}}
     \caption{The NIF values for each question in Perception Test.}
     \label{perception_test_nif3}
     \end{center}
     \vskip -0.3in
\end{figure*}

\begin{figure*}[t]
    % \vskip -0.1in
    \begin{center}
    \centerline{\includegraphics[width=0.55\linewidth]{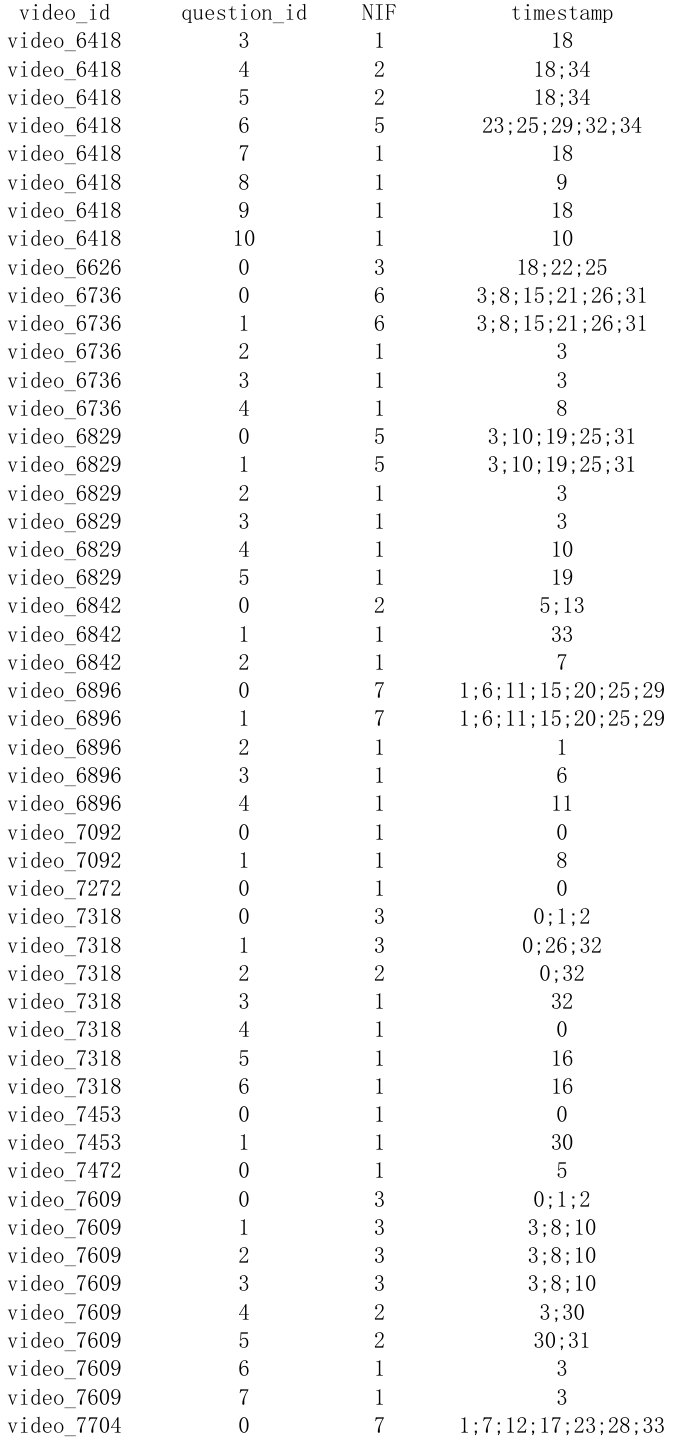}}
     \caption{The NIF values for each question in Perception Test.}
     \label{perception_test_nif4}
     \end{center}
     \vskip -0.3in
\end{figure*}

\begin{figure*}[t]
    % \vskip -0.1in
    \begin{center}
    \centerline{\includegraphics[width=0.55\linewidth]{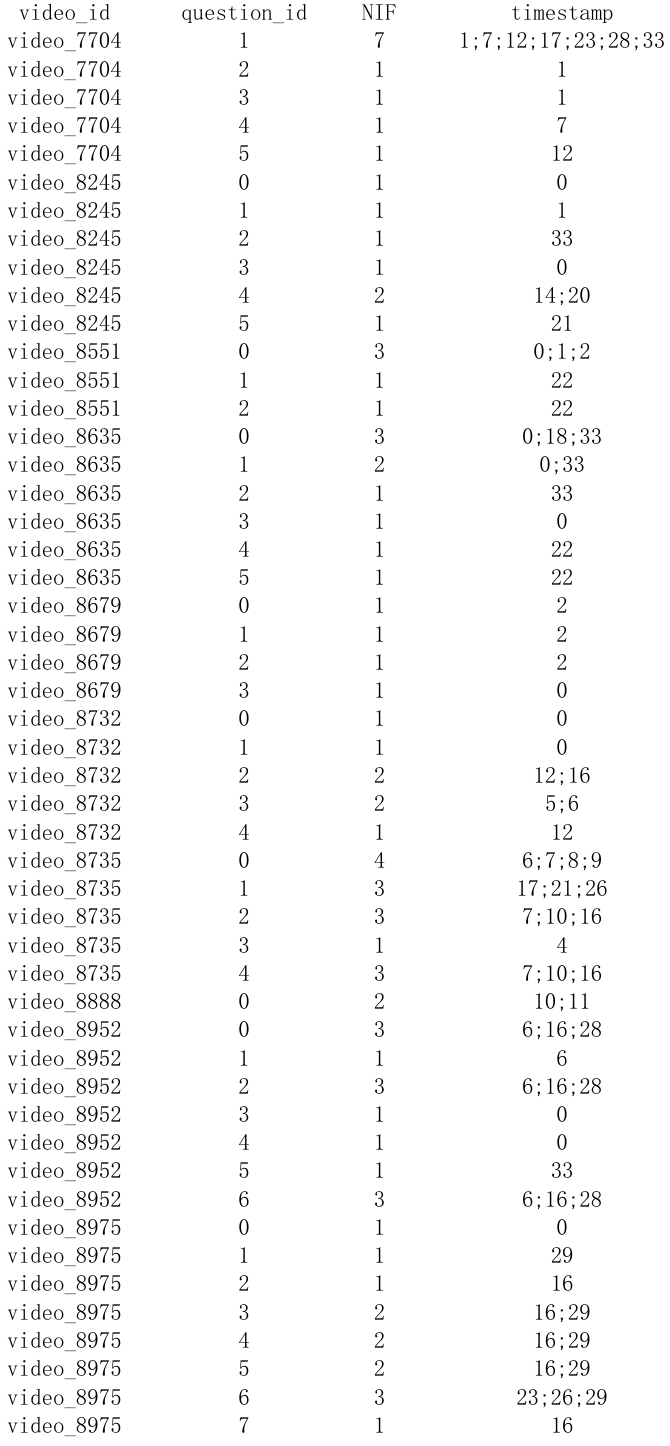}}
     \caption{The NIF values for each question in Perception Test.}
     \label{perception_test_nif5}
     \end{center}
     \vskip -0.3in
\end{figure*}

\begin{figure*}[t]
    % \vskip -0.1in
    \begin{center}
    \centerline{\includegraphics[width=0.55\linewidth]{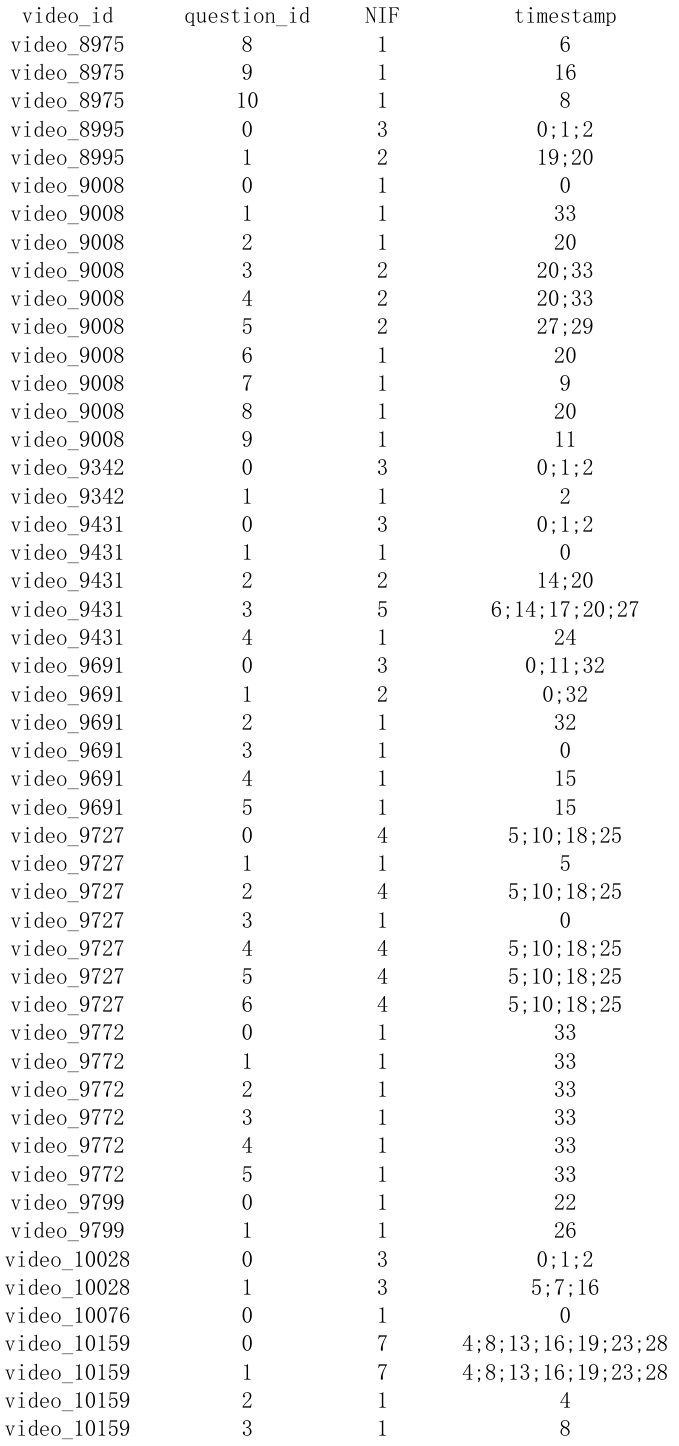}}
     \caption{The NIF values for each question in Perception Test.}
     \label{perception_test_nif6}
     \end{center}
     \vskip -0.3in
\end{figure*}

\begin{figure*}[t]
    % \vskip -0.1in
    \begin{center}
    \centerline{\includegraphics[width=0.55\linewidth]{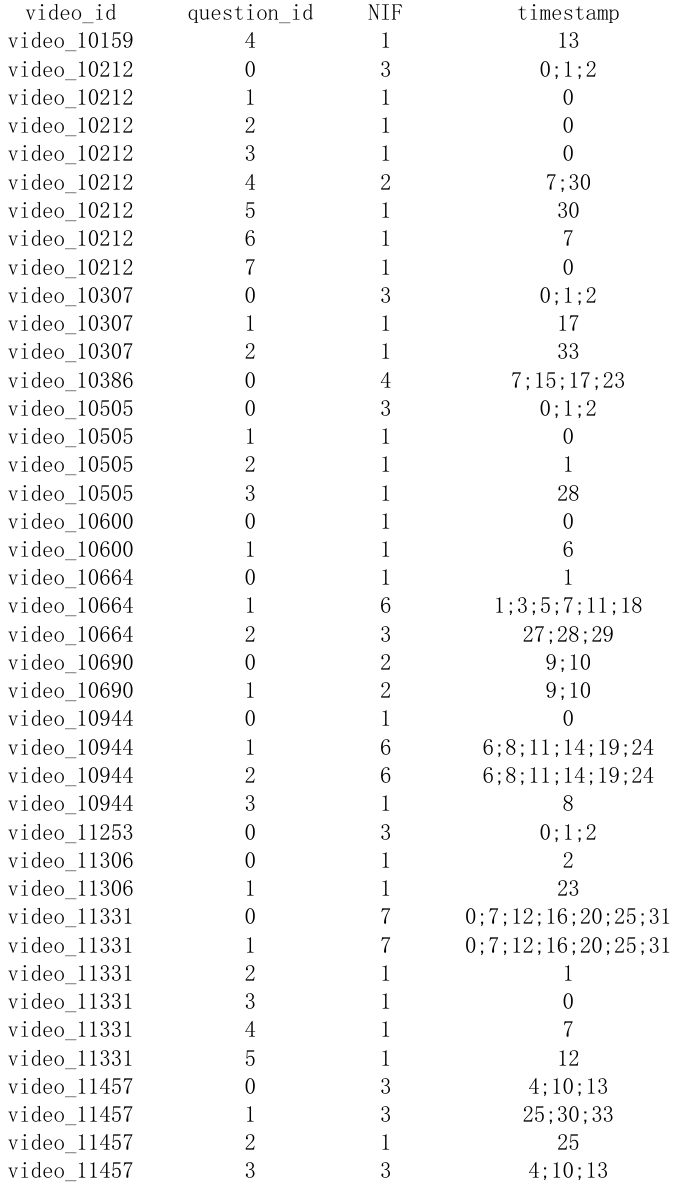}}
     \caption{The NIF values for each question in Perception Test.}
     \label{perception_test_nif7}
     \end{center}
     \vskip -0.3in
\end{figure*}

\begin{figure*}[t]
    % \vskip -0.1in
    \begin{center}
    \centerline{\includegraphics[width=0.55\linewidth]{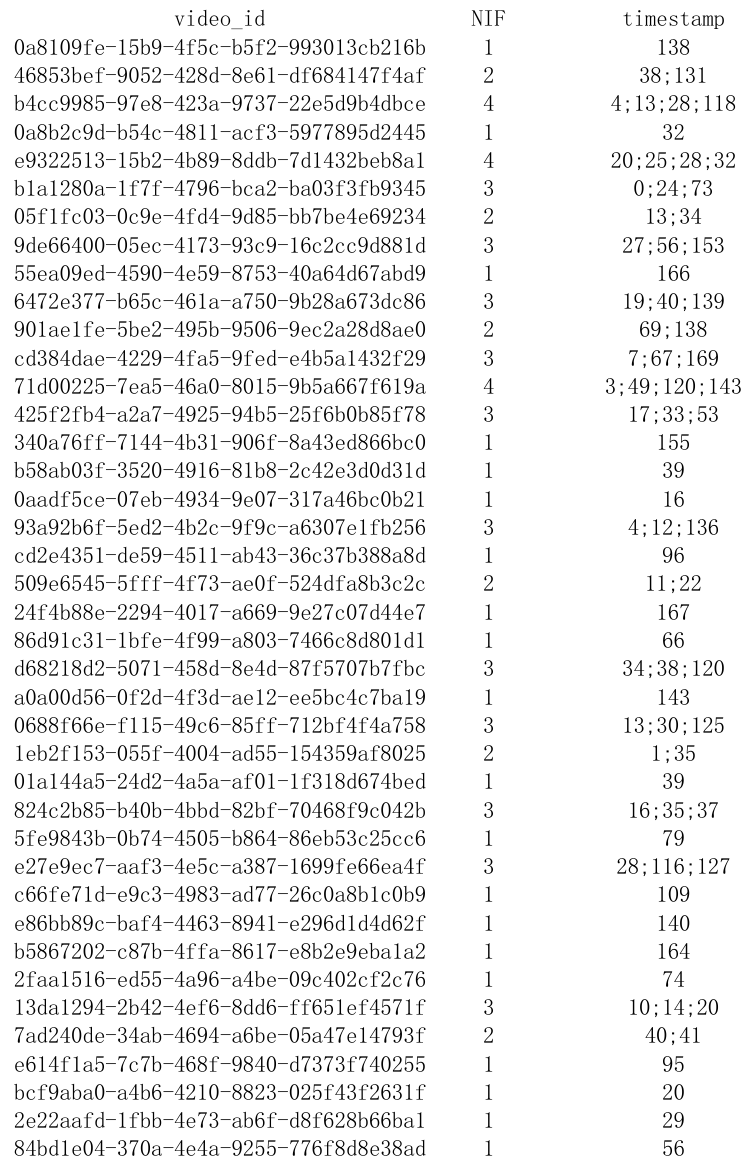}}
     \caption{The NIF values for each question in EgoSchema.}
     \label{egoschema_nif1}
     \end{center}
     \vskip -0.3in
\end{figure*}

\begin{figure*}[t]
    % \vskip -0.1in
    \begin{center}
    \centerline{\includegraphics[width=0.55\linewidth]{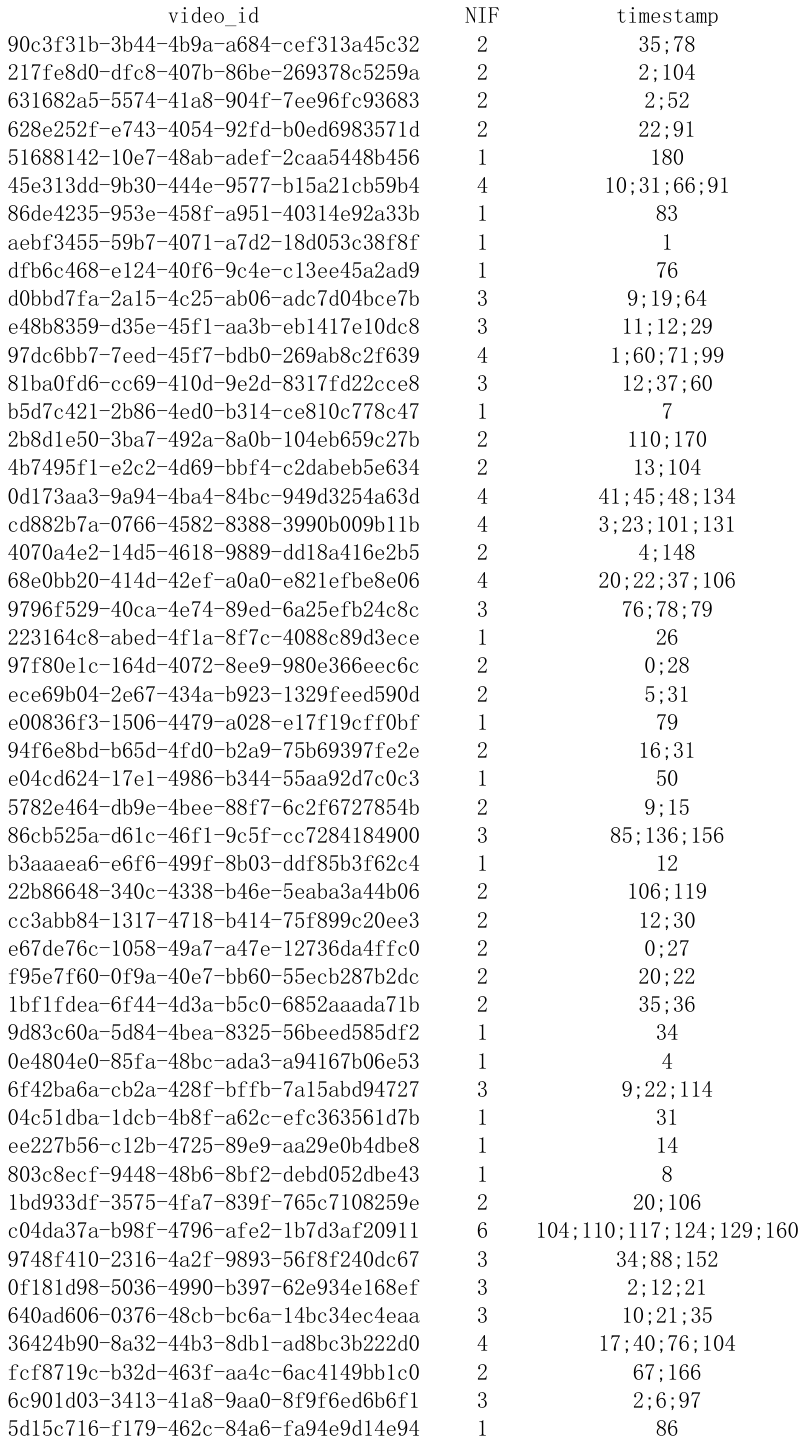}}
     \caption{The NIF values for each question in EgoSchema.}
     \label{egoschema_nif2}
     \end{center}
     \vskip -0.3in
\end{figure*}

In~\Cref{sec:intro}, we propose the concept of the Necessary Information Frame (NIF), which represents the average minimum number of frames containing essential information needed to answer each question.~\Cref{nie} in the paper presents the NIF values for test benchmarks. To better validate the NIF metric, we manually collect the necessary frames (NIF value for each question) and their corresponding timestamps (in seconds) for all questions across four benchmarks. \Cref{video_mme_nif1,video_mme_nif2,video_mme_nif3,video_mme_nif4,video_mme_nif5,video_mme_nif6} illustrate the statistics for Video-MME, including the question IDs and corresponding video IDs. \Cref{mlvu_nif1,mlvu_nif2,mlvu_nif3,mlvu_nif4,mlvu_nif5} show similar data for MLVU, including partial question content, as MLVU lacks specific question IDs, along with the corresponding task and video ID. 
\Cref{perception_test_nif1,perception_test_nif2,perception_test_nif3,perception_test_nif4,perception_test_nif5,perception_test_nif6,perception_test_nif7} and \Cref{egoschema_nif1,egoschema_nif2} correspond to the relevant data for Perception Test and EgoSchema, respectively.

% \newpage

\end{document}